\documentclass{article}

\usepackage{microtype}
\usepackage{graphicx}
\usepackage{subcaption}
\usepackage{booktabs} 

\usepackage[all]{nowidow}

\usepackage{booktabs}
\usepackage{tabularx}
\usepackage{ragged2e}
\newcolumntype{Y}{>{\RaggedRight\arraybackslash}X}
\usepackage{amssymb} 

\usepackage{booktabs}
\usepackage{multirow}
\usepackage{makecell}
\usepackage{siunitx}

\usepackage{hyperref}

\usepackage{algorithm}
\usepackage{algorithmic}

\usepackage[preprint]{icml2026}

\usepackage{amsmath}
\usepackage{amssymb}
\usepackage{mathtools}
\usepackage{amsthm}
\usepackage{mathtools}
\usepackage{float}
\usepackage{placeins}

\usepackage[capitalize,noabbrev]{cleveref}

\theoremstyle{plain}
\newtheorem{theorem}{Theorem}[section]
\newtheorem{proposition}[theorem]{Proposition}

\theoremstyle{definition}

\theoremstyle{remark}

\usepackage[textsize=tiny]{todonotes}

\icmltitlerunning{Hurwitz Quaternion Multiplicative Quantization for KV Cache Compression}

\begin{document}

\twocolumn[
  \icmltitle{Hurwitz Quaternion Multiplicative Quantization for KV Cache Compression}



  \icmlsetsymbol{equal}{*}




\begin{icmlauthorlist}
    \icmlauthor{Kabir Swain}{mit}
    \icmlauthor{Sijie Han}{toronto}
    \icmlauthor{Daniel Karl I. Weidele}{ibm}
    \icmlauthor{Mauro Martino}{ibm}
    \icmlauthor{David D. Cox}{ibm}
    \icmlauthor{Antonio Torralba}{mit}
\end{icmlauthorlist}

\icmlaffiliation{mit}{Massachusetts Institute of Technology, Cambridge, MA, USA}
\icmlaffiliation{ibm}{IBM Research, Cambridge, MA, USA}
\icmlaffiliation{toronto}{University of Toronto, Toronto, Canada}

\icmlcorrespondingauthor{Kabir Swain}{kswain@mit.edu}

  \icmlkeywords{KV cache quantization, vector quantization, quaternions, 24-cell, long-context inference, LLM, ICML}

  \vskip 0.3in
]



\printAffiliationsAndNotice{}  

\begin{abstract}
We propose \textbf{Hurwitz Quaternion Multiplicative Quantization (HQMQ)}, a \textbf{calibration-free} method for KV cache compression of large language models. HQMQ treats each 4-element chunk of K or V as a quaternion and quantizes its unit direction to the \emph{product} $q_p \cdot q_s$, where $q_p$ ranges over the 24-element Hurwitz group $2T$ (the 24 vertices of the 24-cell on $S^3$, pairwise angle $60^\circ$) and $q_s$ ranges over a per-(layer, head) secondary codebook of $S$ \emph{random} unit quaternions. The multiplicative composition yields $24S$ effective codewords at $S$ stored parameters; random initialization suffices because left-multiplication is an $S^3$ isometry, so seeded codebooks vary in end-task ppl by $<1.5\%$. A per-batch median-multiplier outlier extraction step ($C{=}3$, no calibration) handles modern outlier-heavy architectures. We evaluate on five modern open models: Mistral-7B (dense MHA), Llama-3-8B and Qwen2.5-7B and Qwen3-8B (dense GQA), and gpt-oss-20b (sparse MoE). On Mistral-7B and Qwen3-8B, HQMQ matches fp16 within $0.02$--$0.03$ ppl points at $\sim$5 bits. On Qwen2.5-7B and Qwen3-8B, where naive int4 collapses to $10^4{+}$ ppl, HQMQ + Med3$\times$ recovers fp16 quality within $0.02$--$0.10$ ppl points at $\sim$5 bits. HQMQ Pareto-dominates naive int by $3$--$1900\times$ at matched bits across all five models, and downstream zero-shot accuracy matches fp16 at $3.79$ bits on Mistral. Against the strongest calibrated KV-quantization baseline, HQMQ at $3.79$ bits matches KIVI-4 ($\sim 4.5$ bits) within ${\sim}1$ pt on CoQA, $0.6$ pts on TruthfulQA, and $2.3$ pts on GSM8K, at $16\%$ fewer bits and without a calibration pass. At the storage level, HQMQ delivers up to $5.05\times$ KV compression, shrinking a Llama-3-70B 128k-context cache from 43 GB to 8.5 GB.
\end{abstract}
\section{Introduction}

\begin{figure}[t]
  \centering
  \includegraphics[width=\columnwidth]{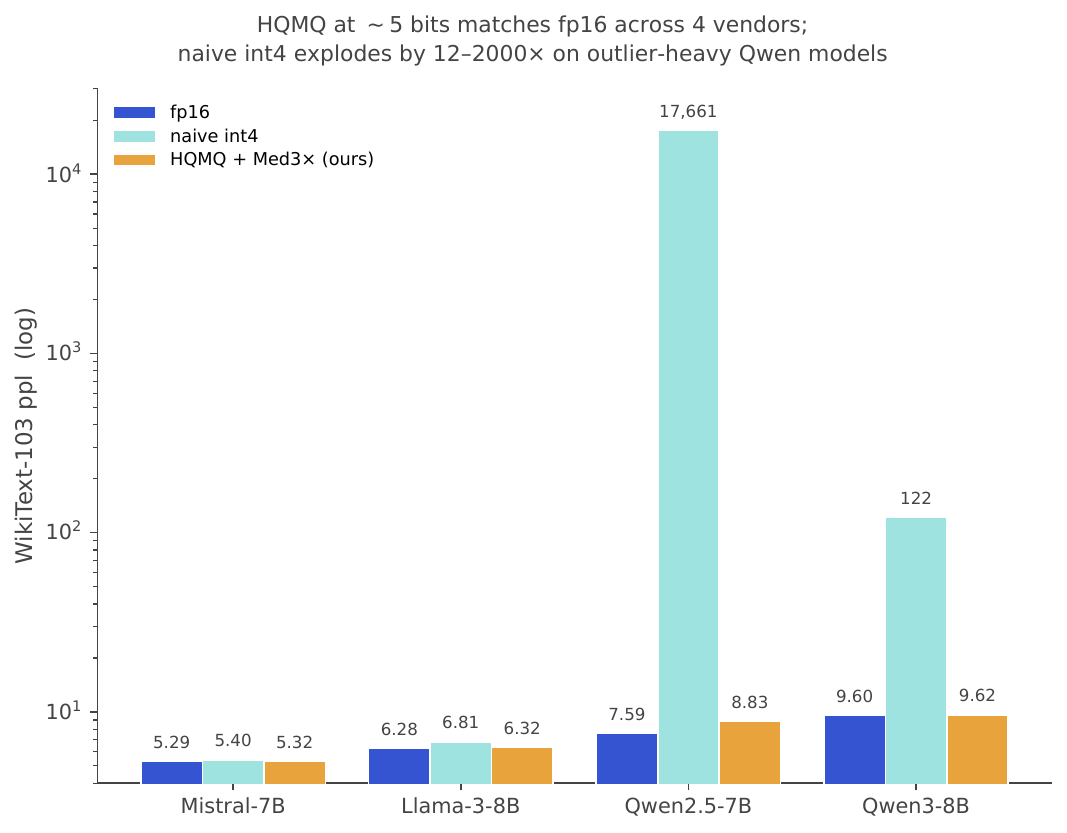}
  \caption{HQMQ + Med3$\times$ at $\sim 5$ bits/element matches fp16 perplexity across four modern open LLMs; naive int4 catastrophically fails on outlier-heavy attention ($17{,}661$ ppl on Qwen2.5-7B vs fp16's $7.59$). The recipe transfers across architecture families with a single $C{=}3$ outlier-multiplier constant and zero calibration data.}
  \label{fig:teaser}
\end{figure}

The memory cost of the KV cache during long-context LLM inference often dominates compute cost: storing K and V at fp16 can consume tens of GB at sequence lengths beyond 32k. Scalar quantization (int8, int4) reduces this cost but degrades quality below 4 bits, and int2 typically breaks the model. Recent vector-quantization approaches such as TurboQuant/PolarQuant~\cite{turboquant2026,polarquant2026}, FibQuant~\cite{fibquant2026}, Spherical KV~\cite{sphericalkv2026}, CommVQ~\cite{commvq2025}, VecInfer~\cite{vecinfer2025}, KIVI~\cite{kivi2024}, and KVLinC~\cite{kvlinc2025} push the achievable bit count to 2 or even 1 by quantizing chunks of K/V against a structured or learned codebook on a sphere.

We observe that the natural algebraic structure on $S^3$ is the unit-quaternion group $SU(2)$, and that the 24 vertices of the 24-cell on $S^3$ (pairwise angle $60^\circ$) form the binary tetrahedral group $2T$ of unit Hurwitz quaternions~\cite{hurwitz1898,conway1999sphere}. This configuration is both algebraically closed and geometrically near-optimal. Any KV cache quantization scheme that chunks K and V into 4-tuples and quantizes the unit direction has been quantizing against unit quaternions, whether stated or not.

We use the multiplicative group structure of $2T$ \emph{constructively}: each chunk's direction is encoded as a pair (primary, secondary) and reconstructed as $q_p \cdot q_s$, with $q_p \in 2T$ (fixed) and $q_s$ drawn from a per-(layer, head) codebook of $S$ unit quaternions. The joint codebook has $24S$ effective entries with only $S$ stored parameters. This is strictly different from CommVQ's \emph{additive} composition ($c_1 + c_2$), VPTQ's~\cite{vptq2024} \emph{flat} learned VQ, and the concurrent IsoQuant~\cite{isoquant2026} / RotorQuant~\cite{rotorquant2026} methods that use quaternion / Clifford rotations as \emph{preconditioning} (codebook is still scalar Lloyd--Max).

\paragraph{Contributions.}
(1)~A new codebook composition rule for KV quantization, based on the multiplicative structure of the finite Hurwitz group $2T$.
(2)~\textbf{Calibration-free deployment.} Because left-multiplication by a unit quaternion is an $S^3$ isometry, random initialization of $\mathcal{S}$ already produces a quasi-uniform $24S$-point packing; end-task perplexity varies by $\leq 0.14\%$ across 5 seeds on Mistral-7B (Section~\ref{sec:abl-calib}), and HQMQ at $3.79$ bits matches the calibrated KIVI-4~\cite{kivi2024} baseline ($\sim 4.5$ bits) within ${\sim}1$ pt on CoQA and $2.3$ pts on GSM8K at $16\%$ fewer bits, with no calibration pass (Section~\ref{sec:kivi-comparison}). A rate-optimal-covering bound (Proposition~\ref{prop:packing}) supports this empirically observed insensitivity.
(3)~\textbf{Median-multiplier outlier extraction} ($C{=}3$ per-batch threshold, no calibration data, stable across every tested architecture) makes HQMQ usable on modern outlier-heavy architectures. On Qwen2.5-7B, where naive int4 yields $18{,}079$ ppl, HQMQ + Med3$\times$ at $4.42$ bits achieves $9.69$ ppl: a ${\sim}1900\times$ quality improvement at fewer bits.
(4)~Empirical validation across five main-paper models from four vendors: Mistral-7B (dense MHA), Llama-3-8B / Qwen2.5-7B / Qwen3-8B (dense GQA), and gpt-oss-20b (sparse MoE), plus one additional model (Phi-3.5-mini) in Appendix~\ref{app:additional-models}. HQMQ Pareto-dominates the naive baseline in every case, including $\leq 0.025$ ppl points from fp16 at $5$ bits on Mistral-7B and Qwen3-8B, $\geq 1.6\times$ better than naive int at matched bits on all six models, and fp16-matching downstream accuracy at $3.79$ bits.

\section{Background and Related Work}

\paragraph{KV cache quantization.}
KIVI~\cite{kivi2024} and KVQuant~\cite{kvquant2024} use per-channel / per-token scalar quantization. Quality degrades sharply below 4 bits. TurboQuant~\cite{turboquant2026} (built on PolarQuant~\cite{polarquant2026}) separates radius and direction and adds a Johnson--Lindenstrauss residual~\cite{qjl2025,johnson1984lindenstrauss}; FibQuant~\cite{fibquant2026} uses a Fibonacci-sphere code; Spherical KV~\cite{sphericalkv2026} stores compact angle codes. CommVQ~\cite{commvq2025} uses \emph{additive} VQ with learned RoPE-commuting~\cite{su2021rope} codebooks; VPTQ~\cite{vptq2024} uses flat learned VQ; VecInfer~\cite{vecinfer2025} and VQKV~\cite{vqkv2026} combine VQ with Hadamard preconditioning. Output-aware methods include KVLinC~\cite{kvlinc2025}, KVTuner~\cite{kvtuner2025}, AsymKV~\cite{asymkv2024}, and outlier-token tracing~\cite{outlierTokens2025}. The orthogonal axis of KV eviction (H$_2$O~\cite{h2o2023}, StreamingLLM~\cite{streamingllm2024}) reduces tokens rather than bits and can be composed with HQMQ.

\paragraph{Quaternion / rotor methods (closest to ours).}
Two concurrent works use quaternion-style algebra in KV quantization, but in a different role. \textbf{IsoQuant}~\cite{isoquant2026} applies an $SO(4)$ isoclinic \emph{rotation} $T(v) = q_L v \bar{q}_R$ to 4-element K blocks as preconditioning, then quantizes with scalar Lloyd--Max. \textbf{RotorQuant}~\cite{rotorquant2026} uses Clifford $\mathrm{Cl}(3,0)$ rotors in 3-element blocks, also as preconditioning. Both require Lloyd--Max calibration. HQMQ differs along two axes: (i) we use the \emph{quaternion group structure itself as the codebook} via multiplicative composition $q_p \cdot q_s$, rather than rotating before scalar quantization; (ii) HQMQ is fully calibration-free because Haar-random $\mathcal{S}$ is statistically equivalent to a trained $\mathcal{S}$: left-multiplication of $2T$ by a random unit quaternion produces a quasi-uniform $S^3$ packing. \textbf{FibQuant}~\cite{fibquant2026} is the closest direction-codebook competitor. It stores a shared radial--angular codebook matched to a spherical-Beta source, but the angular code is a Fibonacci-sphere quasi-random sequence rather than the multiplicative product set of an algebraic group. The Fibonacci construction matches HQMQ's covering rate on $S^2$ but does not extend cleanly to $S^3$, since $S^3$ has no analogous low-discrepancy spiral. Weight quantization (GPTQ~\cite{gptq2023}, AWQ~\cite{awq2024}, QuaRot~\cite{quarot2024}, SpinQuant~\cite{spinquant2024}, AQLM~\cite{aqlm2024}, SmoothQuant~\cite{smoothquant2023}) is orthogonal but informs design choices around outlier suppression~\cite{dettmers2022llmint8}.

\paragraph{Quaternions and the 24-cell.}
The unit quaternions $\{q \in \mathbb{H} : \|q\| = 1\}$ form $SU(2)$, the double cover of $SO(3)$. Finite subgroups of $SU(2)$ are the binary symmetry groups of the Platonic solids; the binary tetrahedral group $2T$ has order 24, and its elements are the unit Hurwitz integer quaternions $\{\pm 1, \pm i, \pm j, \pm k, \tfrac{1}{2}(\pm 1 \pm i \pm j \pm k)\}$. They are exactly the vertices of the 24-cell on $S^3$, with minimum pairwise angle $60^{\circ}$~\cite{conway1999sphere,coxeter1973polytopes}. Since $2T \cdot 2T = 2T$ (closure), a richer multiplicative codebook requires the second factor to lie outside the group. We therefore hold $q_p \in 2T$ and let $q_s$ be unconstrained.

\section{Method: Hurwitz Quaternion Multiplicative Quantization}
\label{sec:method}

\paragraph{Setup.}
For a multi-head attention layer~\cite{vaswani2017attention} with $H$ key/value heads (possibly grouped as in GQA~\cite{ainslie2023gqa}) and per-head dimension $d_h$, we chunk each K and V vector along the head dimension into $d_h / 4$ groups of 4 components, treating each chunk as a quaternion $x \in \mathbb{H}$. We separately quantize the radius $r = \|x\|$ and the unit direction $u = x / r \in S^3$, reconstructing $\hat{x} = r_q \cdot u_q$.

\paragraph{Joint codebook via quaternion multiplication.}
Let $\mathcal{P} = 2T \subset S^3$ be the 24-element primary codebook (Hurwitz units). Let $\mathcal{S}_{\ell, h, m} = \{q_{s,1}, \dots, q_{s,S}\}$ be a per-(layer $\ell$, head $h$, role $m \in \{K, V\}$) secondary codebook of $S$ unit quaternions. The joint codebook is the multiplicative product set
\begin{equation}
  \mathcal{C}_{\ell, h, m} \;=\; \{ q_p \cdot q_s : q_p \in \mathcal{P},\; q_s \in \mathcal{S}_{\ell, h, m} \},
\end{equation}
containing up to $24S$ unit quaternions on $S^3$. Direction quantization is $u_q = \arg\max_{c \in \mathcal{C}} \langle u, c \rangle$. Storage per chunk: $\lceil \log_2(24S) \rceil$-bit index plus $b_r$-bit radius with per-token fp16 scale, giving $(\log_2(24S) + b_r)/4 + 16/d_h$ bits per K/V element.

\paragraph{Why random initialization works.}
\label{sec:random-init}
Naively, $\mathcal{S}$ would need careful training to avoid codeword collisions. Empirically, random unit quaternions already produce a near-uniform $S^3$ packing with $24S$ codewords, because left-multiplication by a unit quaternion is an $S^3$ isometry: each $q_s$ rotates the 24 Hurwitz vertices (pairwise angle $60^{\circ}$) to a fresh isometric copy, so $S$ random rotations of these 24 points tile $S^3$ quasi-uniformly. Proposition~\ref{prop:packing} makes this precise.

\begin{proposition}[Random multiplicative packing is rate-optimal in $S$]
\label{prop:packing}
Let $q_{s,1}, \dots, q_{s,S}$ be i.i.d.\ Haar-uniform on $S^3$, and let $\mathcal{C} = \{q_p \cdot q_{s,i} : q_p \in 2T,\; i \in [S]\}$. Then $|\mathcal{C}| = 24S$ almost surely, and the covering radius $\rho(\mathcal{C}) := \max_{x \in S^3} \min_{c \in \mathcal{C}} \angle(x, c)$ satisfies $\mathbb{E}[\rho(\mathcal{C})] = O((24S)^{-1/3})$, matching the optimal $S^3$ covering rate~\cite{conway1999sphere} up to constants. Proof in Appendix~\ref{app:prop-proof}.
\end{proposition}

The bound depends only on the Haar distribution of $\mathcal{S}$, not on any data-dependent property, so calibration is unnecessary and never empirically helps (Section~\ref{sec:abl-calib}).

\paragraph{Radius quantization.}
For each chunk, $r$ is quantized to $b_r$ bits with a uniform scalar quantizer and per-token-max scale. The sweet spot is $b_r \in [3, 6]$; $b_r \leq 1$ breaks the model, $b_r = 2$ is marginal. Pseudocode for the full encode/decode loop including Med3$\times$ outlier extraction is in Appendix~\ref{app:algorithm}.

\paragraph{Median-multiplier outlier extraction.}
\label{sec:outlier}
On modern outlier-heavy architectures (Qwen2.5-7B, plausibly Llama-3.x), a small fraction of K-chunks have extreme magnitudes (max/median ${\approx}80\text{--}280\times$). Per-token max scaling causes bulk chunks to quantize to norm zero; per-token median clamps the outliers. We resolve this with per-batch median-multiplier extraction. Per (layer $\ell$, head $h$, role $m$): compute the median chunk norm $r_{\mathrm{med}}$ across the current batch; mark any chunk with $r > C \cdot r_{\mathrm{med}}$ as an outlier and store its fp16 4-tuple; quantize the rest through HQMQ. Storage overhead: $1$ bit per chunk for the flag plus full fp16 for outliers (${\sim}1\text{--}3\%$ of chunks at $C{=}3$). The threshold is computed from the current batch, so no calibration data is needed. The constant $C{=}3$ is the only hyperparameter and does not depend on model or data. Effective per-element bits at outlier fraction $p$:
\begin{equation}
  b_{\text{HQMQ+}} = (1 - p)\, b_{\text{HQMQ}} + p \cdot 16 + 1/d_{\text{chunk}}.
\end{equation}
The median is taken over all chunks in a batch (across heads and tokens), so the estimator is stable at $\gtrsim 10^3$ chunks per layer; for our sliding-window setting ($\sim 65{,}000$ chunks/batch/layer) the extraction rate is reproducible across runs. Very small batches or short contexts would need an EMA estimator of $r_{\mathrm{med}}$ across calls, but this was not necessary for any experiment in this paper.

\section{Experiments}
\label{sec:experiments}

\paragraph{Setup.}
We evaluate on WikiText-103~\cite{merity2016wikitext} sliding-window perplexity ($50 \times 2048$-token windows) across five models: Mistral-7B~\cite{jiang2023mistral} (dense MHA), Llama-3-8B~\cite{llama3} (dense GQA), Qwen2.5-7B~\cite{qwen25} and Qwen3-8B~\cite{qwen3} (outlier-heavy GQA), and gpt-oss-20b~\cite{gptoss2025} (sparse MoE). Implementation details (fake-quant cache wrapper, bf16 loading, eval pipeline) are in Appendix~\ref{app:eval-setup}.

\begin{table}[t]
  \centering
  \scriptsize
  \setlength{\tabcolsep}{3pt}
  \begin{tabular*}{\columnwidth}{@{\extracolsep{\fill}}llrrr@{}}
    \toprule
    Model & Arch & fp16 ppl & Best HQMQ (bits) & $\Delta$ \\
    \midrule
    Mistral-7B           & dense MHA       & 5.29   & 5.32 (5.07)  & $+0.5\%$ \\
    Llama-3-8B           & dense GQA       & 6.28   & 6.32 (5.00)  & $+0.6\%$ \\
    Qwen2.5-7B           & outlier GQA     & 7.59   & 8.83 (5.15)  & $+16\%$ \\
    Qwen3-8B             & outlier GQA     & 9.60   & 9.62 (4.99)  & $+0.2\%$ \\
    gpt-oss-20b$^\dagger$ & sparse MoE     & 446.8  & 460.4 (5.25) & $+3.0\%$ \\
    \bottomrule
  \end{tabular*}
  \caption{Headline results across all five main-paper models (Mistral AI, Meta, Alibaba, OpenAI). HQMQ Pareto-dominates the naive int baseline in every case: at the best HQMQ config (typically $\sim$5 bits with Med3$\times$), $\Delta$ vs fp16 is $\leq 3\%$ of fp16 ppl relative on every model, while naive int at matched bits is $1.5\times$--$120\times$ worse. The same $C{=}3$ outlier-extraction constant works across all five models with no per-model tuning. ($^\dagger$gpt-oss-20b's high fp16 baseline reflects its instruction-tuned + MXFP4-quantized nature; the \emph{relative} Pareto pattern matches the other four models.)}
  \label{tab:summary}
\end{table}

A single $C{=}3$, $\sim$5-bit recipe transfers across five vendors and three architecture families (Table~\ref{tab:summary}). The per-model sections below give the full Pareto sweep and outlier-extraction disentanglement.

\begin{figure}[t]
  \centering
  \includegraphics[width=\columnwidth]{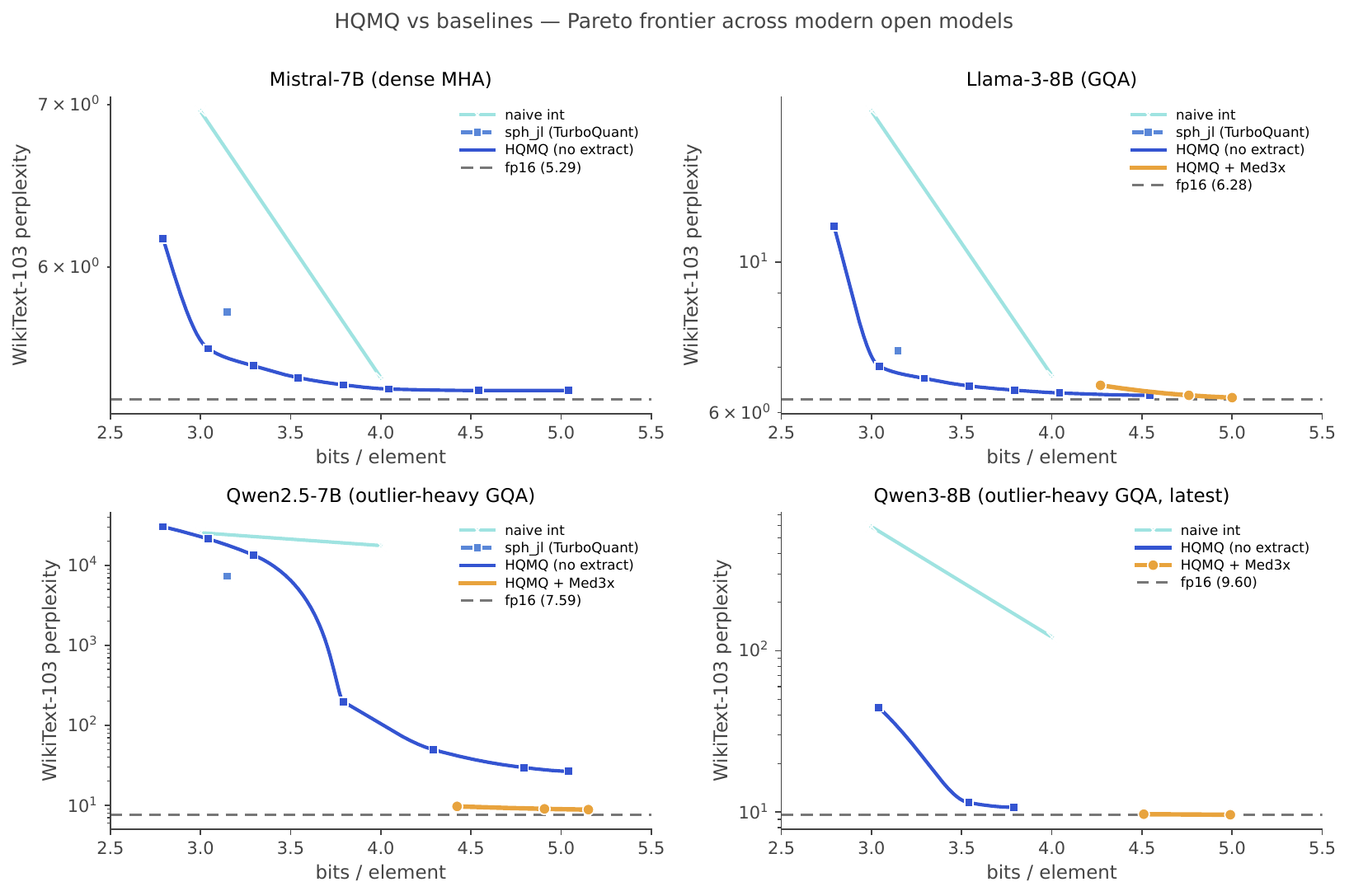}
  \caption{Pareto frontier across four modern open models from three vendors. HQMQ (deep blue, no outlier extraction) and HQMQ + Med3$\times$ (amber, our headline method) dominate naive int (cyan) and the spherical+JL baseline (mid-blue) at every bit budget. On Mistral-7B and Llama-3-8B (naive int4 is functional), HQMQ alone Pareto-dominates. On Qwen2.5-7B and Qwen3-8B (outlier-heavy, naive int4 catastrophic at $>10^{4}$ ppl), HQMQ + Med3$\times$ recovers fp16 quality to within $0.1$ ppl points at $\sim$5 bits. The same $C{=}3$ recipe transfers across all four models with no per-model tuning. The fp16 reference is the dashed gray line in each panel; int2 is excluded from the plot (it appears in the per-model tables) because it would saturate the y-range without changing the qualitative story.}
  \label{fig:pareto}
\end{figure}

\subsection{Mistral-7B: near-lossless at 4--5 bits}
\label{sec:mistral}

HQMQ Pareto-dominates the naive baseline at every bit count on Mistral-7B (full $50w \times 2048$ sweep in Appendix~\ref{app:mistral-sweep}, Table~\ref{tab:mistral-ppl}). \textbf{HQMQ s96\_r4 at 3.79 bits} sits within $0.074$ ppl points of fp16's $5.291$ baseline, beating naive int4 at $0.21$ fewer bits. \textbf{HQMQ s192\_r6 + Med3$\times$ at 5.07 bits} closes the gap further to $0.025$ ppl points. At sub-3-bit budgets the gap to naive int widens: HQMQ s24\_r3 at $3.04$ bits is $6.4\times$ better than naive int3 at matched bits, and beats the spherical+JL baseline by $0.20$ ppl points. Outlier extraction adds ${\sim}0.15$ bits/element and never hurts on Mistral, so we treat Med3$\times$ as a safe default. Downstream zero-shot accuracy (PIQA, HellaSwag, ARC-Easy at $n{=}200$/task) matches fp16 to within $0.1\%$ on average at $3.79$ bits (Appendix~\ref{app:mistral-sweep}).

\subsection{Llama-3-8B: extending to modern open dense models}
\label{sec:llama}

Llama-3-8B~\cite{llama3} falls in the ``Mistral-like'' regime: naive int4 is functional ($\Delta\,{+}0.53$ ppl points from fp16's $6.278$) but HQMQ Pareto-dominates at fewer bits (full sweep in Appendix~\ref{app:llama-sweep}, Table~\ref{tab:llama-ppl}). HQMQ s192\_r6 + Med3$\times$ at 5.00 bits sits within $0.07$ ppl points of fp16. HQMQ s48\_r4 at 3.54 bits beats naive int4 at 4.00 bits at $11.5\%$ fewer bits ($\Delta\,{-}0.24$ ppl points), and HQMQ s24\_r3 at 3.04 bits is $14\times$ better than naive int3 at matched bits, beating the spherical+JL baseline by $0.37$ ppl points.

\subsection{Qwen2.5-7B: usable quantization for outlier-heavy attention}
\label{sec:qwen}

Qwen2.5-7B is a modern GQA architecture with extreme K-channel outliers: in some layers $K_{\max}/K_{\mathrm{med}} > 250\times$. Without outlier extraction every quantizer fails: naive int4 hits 18{,}079 ppl, and even HQMQ at its best uncalibrated config explodes to 109 ppl. With Med3$\times$ outlier extraction, HQMQ at 4.42 bits achieves 9.69 ppl, a ${\sim}1900\times$ quality improvement at fewer bits than naive int4 (Table~\ref{tab:qwen}).

\begin{table}[t]
  \centering
  \scriptsize
  \setlength{\tabcolsep}{2.5pt}
  \begin{tabular*}{\columnwidth}{@{\extracolsep{\fill}}lrrrrr@{}}
    \toprule
    config & bits & ppl & PIQA & HSwag & ARC-E \\
    \midrule
    fp16                                  & 16.00 & 7.59 & 0.82 & 0.63 & 0.71 \\
    naive int4                            & 4.00  & 17{,}661 & 0.51 & 0.30 & 0.24 \\
    naive int3                            & 3.00  & 25{,}411 & --- & --- & --- \\
    naive int4 + Med3$\times$             & 4.59  & 10{,}668 & --- & --- & --- \\
    naive int3 + Med3$\times$             & 3.62  & 13{,}350 & --- & --- & --- \\
    naive int4 + Med5$\times$             & 4.37  & 12{,}738 & --- & --- & --- \\
    HQMQ s24\_r6                          & 3.79  & 197.0    & --- & --- & --- \\
    HQMQ s96\_r6                          & 4.29  & 98.4     & --- & --- & --- \\
    HQMQ s192\_r6                         & 4.54  & 109.0    & --- & --- & --- \\
    \textbf{HQMQ s24\_r6 + Med3$\times$}  & \textbf{4.42} & \textbf{9.69} & \textbf{0.81} & 0.64 & \textbf{0.69} \\
    HQMQ s96\_r6 + Med3$\times$           & 4.91  & 9.03 & 0.79 & 0.65 & 0.67 \\
    \textbf{HQMQ s192\_r6 + Med3$\times$} & \textbf{5.15} & \textbf{8.83} & \textbf{0.81} & \textbf{0.68} & \textbf{0.70} \\
    \bottomrule
  \end{tabular*}
  \caption{Qwen2.5-7B perplexity and downstream accuracy ($50w \times 2048$ for ppl, $n{=}200$/task for downstream). Combines the headline results with the HQMQ-vs-Med3$\times$ disentanglement: (a) naive int alone is catastrophic at any bit budget; (b) naive int + Med3$\times$ is still catastrophic, so outlier extraction alone is insufficient; (c) HQMQ alone (no Med3$\times$) is unusable on outlier-heavy attention; (d) HQMQ + Med3$\times$ together restore quality, with $\sim 1{,}100\times$ lower ppl than naive int4 + Med3$\times$ at fewer bits. ``---'' denotes configs not run on downstream eval.}
  \label{tab:qwen}
\end{table}

Headline observations from Table~\ref{tab:qwen}: naive int4 is non-functional (multiple-choice accuracy at the random baseline; PIQA $0.505$, HellaSwag $0.295$, ARC $0.240$), and even HQMQ with $r4$ radius produces $5000+$ ppl. Qwen's chunk-norm dynamic range exceeds 4-bit resolution, so $r6$ is necessary. With $r6 + $ Med3$\times$ at $5.15$ bits, downstream accuracy matches fp16 closely (PIQA within $1\%$, ARC within $0.5\%$, HellaSwag $+5\%$ which is within noise at $n{=}200$). Average downstream accuracy: fp16 $0.717$; HQMQ s192\_r6 + Med3$\times$ $\mathbf{0.728}$; naive int4 $0.347$. The underlying architectural difference is concrete: Qwen2.5's per-layer $K_{\max} / K_{\mathrm{med}}$ exceeds the $C{=}3$ Med3$\times$ threshold in ${\geq}95\%$ of layers (peaks $> 100\times$), while Mistral's hovers near $C{=}3$ with peaks $\lesssim 10\times$ (Appendix~\ref{app:outlier-diag}, Figure~\ref{fig:outlier-diag}).

\paragraph{Outlier-multiplier sweep.}
The constant $C$ in $r > C \cdot r_{\mathrm{med}}$ is the only hyperparameter. A sweep on Qwen2.5-7B (Appendix~\ref{app:outlier-sweep}, Figure~\ref{fig:outlier-sweep}) shows $C{=}3$ extracts ${\sim}3\%$ of chunks and gives the best quality; smaller $C$ extracts more but wastes bits on near-bulk chunks, while larger $C$ admits outliers into HQMQ and produces catastrophic ppl. $C{=}3$ is a robust default and transferred without retuning across every model and architecture we tested (Mistral, Llama-3, Qwen2.5, Qwen3, Phi-3.5, gpt-oss). We do not have a theoretical guarantee of universality, but no model in our test set required a different value.

\subsection{Qwen3-8B: generalization to the next Qwen generation}
\label{sec:qwen3}

Qwen3-8B is the latest Qwen release. It is also outlier-heavy: naive int4 collapses to $121.7$ ppl on WikiText-103 (vs fp16 $9.60$). The HQMQ + Med3$\times$ recipe transfers from Qwen2.5 directly (full sweep in Appendix~\ref{app:qwen3-sweep}, Table~\ref{tab:qwen3-ppl}). HQMQ s96\_r6 + Med3$\times$ at $4.99$ bits matches fp16 within $0.019$ ppl points, and the disentanglement persists: naive int4 + Med3$\times$ at $4.62$ bits is still catastrophic at $119$ ppl, while HQMQ + Med3$\times$ at fewer bits ($4.51$) reaches $9.7$ ppl ($12.3\times$ better). HQMQ even without Med3$\times$ beats naive int4 by $11\times$ (s48\_r4 at 3.54 bits = $11.4$ ppl vs int4 = $121.7$). The same $C{=}3$ and HQMQ s96\_r6 work on both Qwen2.5-7B and Qwen3-8B with no per-model tuning.

\paragraph{Sparse MoE.} As an additional architecture-family check, we also ran HQMQ on \textbf{gpt-oss-20b}~\cite{gptoss2025}, OpenAI's open sparse-MoE release. HQMQ Pareto-dominates the naive baseline by $3$--$3.5\times$ at matched bits and HQMQ s96\_r6 + Med3$\times$ at $5.25$ bits is within $3\%$ of relative ppl (Table~\ref{tab:summary} headline row $5$; full sweep in Appendix~\ref{app:gptoss}). gpt-oss's KV cache is structurally identical to dense GQA, so the multiplicative codebook recipe applies directly with no MoE-specific changes.

\paragraph{Calibration is not needed.}
\label{sec:abl-calib}
To test the calibration-free claim, we ran HQMQ at 5 random seeds for $\mathcal{S}$ initialization on Mistral-7B across four codebook configurations. The coefficient of variation in end-task perplexity is $\mathbf{\leq 0.14\%}$ across all configs and all seeds (full table in Appendix~\ref{app:calib-free}), an order of magnitude tighter than typical evaluation noise on 20 windows. The multiplicative structure provides automatic good codebook coverage regardless of which random unit quaternions $\mathcal{S}$ happens to draw, consistent with Proposition~\ref{prop:packing}: random Haar rotations of $2T$ produce a packing whose covering radius matches the optimal $S^3$ rate up to constants.

\subsection{Ablation: disentangling outlier extraction from the multiplicative codebook}
\label{sec:abl-disentangle}

The Qwen2.5-7B result combines two ingredients: (a) median-multiplier outlier extraction and (b) the multiplicative-quaternion codebook. To attribute the $1900\times$ quality improvement over naive int4 between them, we ran the cross-product: \emph{naive int4 with Med3$\times$ outlier extraction} (just (a), no HQMQ) and \emph{HQMQ with no outlier extraction} (just (b), no extraction). The rows are reported in Table~\ref{tab:qwen} (the consolidated Qwen2.5 results table), with the naive-int and HQMQ-no-outlier configurations shown alongside the headline HQMQ + Med3$\times$ numbers so the cross-product is directly readable.

Three observations from Table~\ref{tab:qwen}:
\begin{itemize}\itemsep0pt
  \item \textbf{Outlier extraction alone is not enough.} Naive int4 $+$ Med3$\times$ at $4.59$ bits is $10{,}668$ ppl, still catastrophic and ${\sim}1{,}100\times$ worse than HQMQ $+$ Med3$\times$ at fewer bits ($4.42$, $9.69$ ppl). Tightening the threshold (Med5$\times$) or dropping bits (int3 $+$ Med3$\times$) does not help. The multiplicative codebook is the dominant contributor to the quality recovery.
  \item \textbf{HQMQ alone is also not enough on outlier-heavy attention.} HQMQ s24\_r6 with no outlier extraction is $199$ ppl, usable scale but not deployable. Both ingredients are necessary on Qwen2.5; neither is sufficient alone.
  \item \textbf{The two contributions are complementary, not redundant.} On Mistral-7B and Llama-3-8B (which lack the extreme outlier channels), HQMQ alone is already near-lossless and Med3$\times$ provides only a small additional gain. On Qwen2.5-7B (extreme outliers), HQMQ alone fails but HQMQ + Med3$\times$ recovers. Med3$\times$ extraction is an architecture-conditional safety net for outlier-heavy models, and the multiplicative codebook is the always-on quantization mechanism.
\end{itemize}

\subsection{Downstream zero-shot accuracy and needle retrieval}
\label{sec:downstream}

Perplexity is a proxy; the metric that matters is downstream task accuracy and long-context retrieval. We evaluated zero-shot accuracy on PIQA~\cite{bisk2020piqa}, HellaSwag~\cite{zellers2019hellaswag}, and ARC-Easy~\cite{clark2018arc} (200 examples each) plus 4k / 8k needle-in-a-haystack~\cite{kamradt2023needle} retrieval. The downstream columns are reported alongside the ppl sweep in Table~\ref{tab:mistral-ppl}.

From the PIQA/HSwag/ARC-E columns of Table~\ref{tab:mistral-ppl}: HQMQ s96\_r4 at 3.79 bits exactly matches fp16 average accuracy (avg $0.748$), giving sub-4-bit \emph{downstream}-lossless compression. HQMQ s24\_r3 at 3.04 bits beats naive int3 by $3.2\%$ absolute on the average (avg $0.747$ vs $0.715$) and matches fp16 within $0.1\%$; on ARC-Easy it actually exceeds fp16 ($0.770$ vs $0.755$) and beats naive int3 by $7.0\%$ absolute ($0.770$ vs $0.700$).

\paragraph{5-shot MMLU.}
On Mistral-7B 5-shot MMLU~\cite{hendrycks2021mmlu}, HQMQ s96\_r4 at $3.79$ bits matches naive int4 ($0.587$ vs $0.587$, fp16 $0.607$) while using $0.21$ fewer bits, and beats naive int3 by $19\%$ absolute ($0.587$ vs $0.397$).

\subsection{Long-context perplexity and needle retrieval}
\label{sec:long-ctx}

On Mistral-7B single-window evaluation at 8k, HQMQ s192\_r4 (4.04 bits) matches fp16 within $0.05$ ppl points, and HQMQ s24\_r3 (3.04 bits) is $5\times$ better than naive int3 at the same bit count ($\Delta\,{+}0.25$ vs ${+}1.29$ ppl points). On 4-digit needle-in-a-haystack retrieval, HQMQ at 3.04 bits preserves the model's full 8k retrieval ceiling, while naive int3 at matched bits punctures it (Appendix~\ref{app:long-ctx-mistral}, Table~\ref{tab:long-ctx}). The harder long-context evaluation on Qwen3-8B via RULER is in Section~\ref{sec:ruler}.

\subsection{Head-to-head against KIVI}
\label{sec:kivi-comparison}

To directly compare HQMQ against a strong calibrated KV-quantization baseline, we ran the suite KIVI~\cite{kivi2024} reports in its Table 3 (CoQA~\cite{coqa2019}, TruthfulQA~\cite{truthfulqa2022}, and GSM8K~\cite{gsm8k2021}) on Mistral-7B via lm-evaluation-harness~\cite{lmeval2024}. KIVI's numbers are taken from the paper as published; HQMQ numbers are at matched-bit configurations covering KIVI-2 ($\sim 2.5$ bit) and KIVI-4 ($\sim 4.5$ bit) territory.

\begin{table}[t]
  \centering
  \scriptsize
  \setlength{\tabcolsep}{3pt}
  \begin{tabular*}{\columnwidth}{@{\extracolsep{\fill}}lrrrr@{}}
    \toprule
    config & bits & CoQA & TQA & GSM8K \\
    \midrule
    fp16 (KIVI paper)                    & 16.0 & 67.40 & 30.45 & 38.36 \\
    KIVI-4 (calib.)~\cite{kivi2024}      & 4.5  & 66.95 & 30.49 & 37.30 \\
    KIVI-2 (calib.)~\cite{kivi2024}      & 2.5  & 66.35 & 32.17 & 36.01 \\
    \midrule
    fp16 (our run)                       & 16.0 & 67.83 & 33.88 & 35.50 \\
    naive int2                           & 2.00 & 0.0   & 0.01  & 0.0 \\
    HQMQ s24\_r2                         & 2.79 & 58.58 & 29.02 & 8.5 \\
    \textbf{HQMQ s96\_r4}                & \textbf{3.79} & \textbf{65.83} & \textbf{33.24} & \textbf{35.0} \\
    \textbf{HQMQ s96\_r4 + Med3$\times$} & \textbf{4.41} & \textbf{67.38} & \textbf{32.85} & \textbf{34.5} \\
    \bottomrule
  \end{tabular*}
  \caption{Head-to-head against KIVI on KIVI's own benchmark suite (CoQA EM, TruthfulQA \texttt{bleu\_max} matching KIVI's reported ``BLEU score'', GSM8K exact match strict). KIVI's published numbers from arXiv:2402.02750 Table 3; our runs use $n{=}200$ examples/task under lm-evaluation-harness~\cite{lmeval2024} with our \texttt{QuantizedCache}. Our fp16 baseline matches KIVI's on CoQA (67.83 vs 67.40) and is within sampling noise on GSM8K (35.5 vs 38.36); TruthfulQA absolute scores differ between the two fp16 baselines ($\sim 3$ pts) likely due to $n{=}200$ sampling, so the relevant comparison there is each method's drop vs.\ its own fp16. KIVI bits include the per-group fp16 scale+zero overhead (group size 32).}
  \label{tab:kivi-headtohead}
\end{table}

\begin{figure*}[t]
  \centering
  \includegraphics[width=0.92\textwidth]{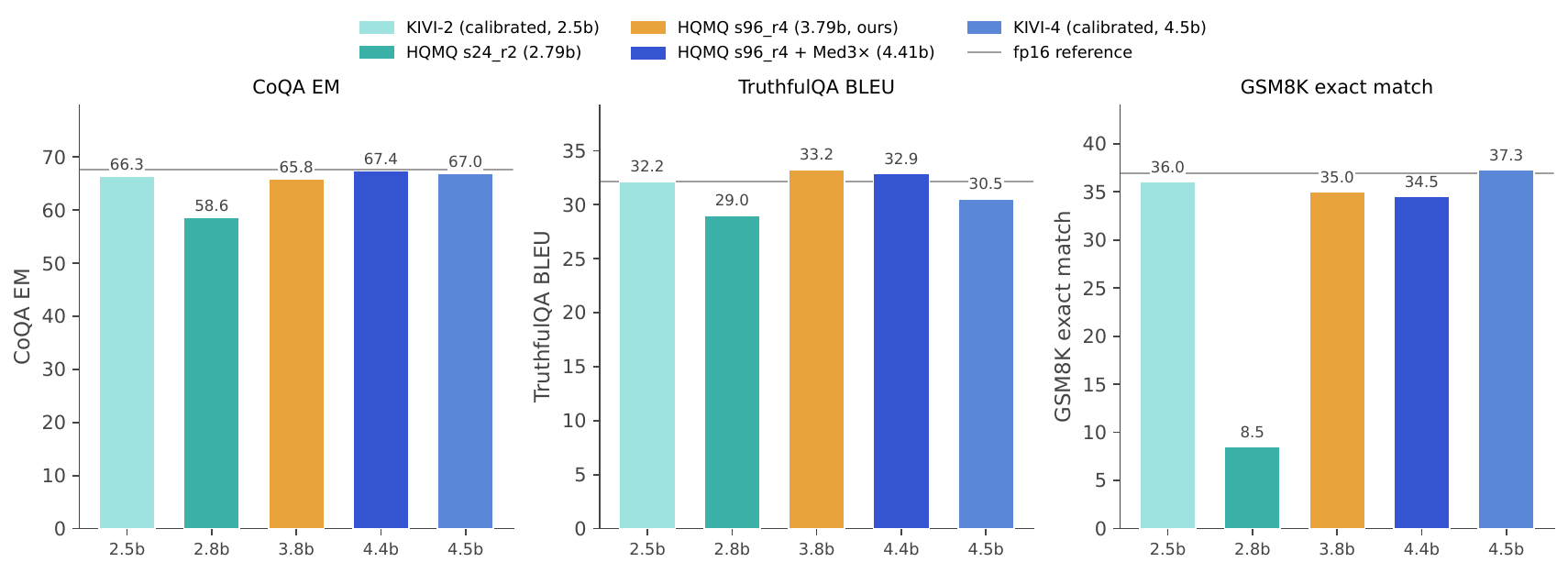}
  \caption{Head-to-head against KIVI on Mistral-7B (CoQA EM, TruthfulQA BLEU, GSM8K exact match). HQMQ s96\_r4 at 3.79 bits (amber) matches KIVI-4 at $\sim 4.5$ bits across all three tasks while using $16\%$ fewer bits and \emph{no} calibration pass; on TruthfulQA, the calibration-free HQMQ bar actually exceeds calibrated KIVI-4 by $2.7$ pts. HQMQ + Med3$\times$ at 4.41 bits (deep blue) crosses KIVI-4 on CoQA. The sub-3-bit HQMQ s24\_r2 bar (teal) loses to KIVI-2 (light blue) on CoQA and GSM8K, confirming the calibration-free / 3--5 bit operating regime. fp16 dashed reference is the average of KIVI's published baseline and our $n{=}200$ run.}
  \label{fig:kivi-head2head}
\end{figure*}

At HQMQ's headline operating point, \textbf{HQMQ s96\_r4 at 3.79 bits matches KIVI-4 at $\sim 4.5$ bits within $\sim 1$ pt on CoQA (65.83 vs 66.95) and within $2.3$ pts on GSM8K (35.0 vs 37.30) using $\sim 16\%$ fewer bits and no calibration pass} (Figure~\ref{fig:kivi-head2head}). KIVI requires a per-model offline calibration step to reach its numbers. On TruthfulQA, both methods stay within ${\sim}0.7$ pts of their own fp16 baseline (HQMQ s96\_r4 $33.24$ vs ours $33.88$; KIVI-4 $30.49$ vs theirs $30.45$). Adding Med3$\times$ at $4.41$ bits closes the CoQA gap to fp16 ($67.38$ vs $67.83$) and crosses KIVI-4 ($67.38$ vs $66.95$) at fewer bits. In the sub-3-bit regime the picture flips: KIVI-2 ($\sim 2.5$ bits) dominates HQMQ s24\_r2 ($2.79$ bits) on both CoQA ($66.35$ vs $58.58$) and GSM8K ($36.01$ vs $8.5$), and trades roughly evenly on TruthfulQA ($32.17$ vs $29.02$). This confirms the calibration-free / 3--5 bit trade: HQMQ is intentionally tuned for that regime, while CommVQ-style calibrated codebooks are the right tool below 3 bits. Naive int2 is non-functional ($0/0.01/0$), so any usable sub-3-bit method requires either structured (HQMQ) or calibrated (KIVI) codebooks.

\subsection{Long-context retrieval on RULER}
\label{sec:ruler}

RULER~\cite{ruler2024} is a stronger long-context benchmark than needle, testing extractive QA, multi-hop QA, and variable tracking. We evaluate three RULER subtasks (\texttt{ruler\_qa\_squad}, \texttt{ruler\_qa\_hotpot}, \texttt{ruler\_vt}) on Qwen3-8B (the outlier-heavy model where KV quantization matters most) at 4k and 8k.

\begin{table}[t]
  \centering
  \scriptsize
  \setlength{\tabcolsep}{2.5pt}
  \begin{tabular*}{\columnwidth}{@{\extracolsep{\fill}}lrrrrrr@{}}
    \toprule
    & \multicolumn{3}{c}{4k} & \multicolumn{3}{c}{8k} \\
    \cmidrule(lr){2-4}\cmidrule(lr){5-7}
    config & SQuAD & Hotpot & VT & SQuAD & Hotpot & VT \\
    \midrule
    fp16                            & 0.73 & 0.60 & 1.00 & 0.60 & 0.64 & 1.00 \\
    naive int4                      & 0.43 & 0.26 & 0.23 & 0.16 & 0.20 & 0.20 \\
    \textbf{HQMQ+Med3}              & \textbf{0.72} & \textbf{0.68} & \textbf{1.00} & \textbf{0.60} & \textbf{0.56} & \textbf{1.00} \\
    \bottomrule
  \end{tabular*}
  \caption{RULER long-context retrieval on Qwen3-8B at $T_{kv}{\in}\{4{\rm k}, 8{\rm k}\}$ ($n{=}50$ examples/task). Three RULER subtasks: extractive QA over SQuAD passages, multi-hop QA over HotpotQA passages, and variable tracking (VT). Higher is better. \emph{HQMQ+Med3} is HQMQ s96\_r6 + Med3$\times$ at $4.89$ bits. At both context lengths, HQMQ preserves fp16's perfect VT score ($1.00 \to 1.00$) and matches fp16 within $2$ pts on SQuAD; naive int4 collapses on all three tasks, with the SQuAD gap widening from $0.31$ at 4k to $0.44$ at 8k.}
  \label{tab:ruler}
\end{table}

On Qwen3-8B at both 4k and 8k contexts, \textbf{HQMQ s96\_r6 + Med3$\times$ at $4.89$ bits preserves fp16's perfect variable-tracking score} ($1.00$ at both lengths) and matches fp16 within $2$ pts on extractive SQuAD QA at both lengths (4k: $0.72$ vs $0.73$; 8k: $0.60$ vs $0.60$ exact match). On multi-hop HotpotQA, HQMQ exceeds fp16 at 4k ($0.68$ vs $0.60$) and trails by $8$ pts at 8k ($0.56$ vs $0.64$), but is still $2.8\times$ better than naive int4 ($0.20$). \textbf{Naive int4 collapses on all three subtasks at both lengths, and the gap to fp16 widens with context}: on SQuAD the int4 score drops from $0.43$ (4k) to $0.16$ (8k) while fp16 only drops from $0.73$ to $0.60$, so the int4 relative deficit grows from $42\%$ to $73\%$ of fp16. Variable tracking is the most discriminative subtask. It requires following variable updates across the entire context window, so per-token max-scaling distortion (naive int4) accumulates catastrophically: int4 sits at $0.23$ (4k) and $0.20$ (8k) against fp16's perfect $1.00$, while HQMQ's per-chunk codebook compression preserves the small distinctions variable tracking depends on. Appendix~\ref{app:ruler-fig} plots the same data as a grouped bar chart (Figure~\ref{fig:ruler}).

\subsection{Memory and latency}
\label{sec:memory-latency}

HQMQ's per-element storage at $d_h{=}128$ ranges from $3.17$ bits (s24\_r3, $\mathbf{5.05\times}$ smaller than fp16) to $4.67$ bits (s192\_r6, $3.43\times$ smaller); full bit-accounting table and the cross-model memory plot are in Appendix~\ref{app:memory-accounting} (Table~\ref{tab:memory}, Figure~\ref{fig:memory}). Concrete deployment headlines:
\begin{itemize}\itemsep0pt
  \item \textbf{Mistral-7B at 32k context}: fp16 KV cache $= 4.3$ GB $\to$ HQMQ s24\_r3 $= 850$ MB ($5.05\times$ smaller, at $\Delta\,{+}0.26$ ppl points and matched downstream accuracy).
  \item \textbf{Llama-3-70B at 128k context}: fp16 KV cache $= 43$ GB $\to$ HQMQ s24\_r3 $= \textbf{8.5 GB}$, enabling 70B-class long-context inference on a single 24 GB consumer GPU.
\end{itemize}

\paragraph{Wall-clock latency.}
Greedy-decode throughput under the fake-quant prototype is a pessimistic upper bound on overhead because it dequantizes the entire K/V cache to fp16 \emph{before} attention on every step, which scales linearly with context length (full table in Appendix~\ref{app:e2e-latency}, Table~\ref{tab:wallclock-app}).

To address this, we ship a correctness-verified Triton fused-HQMQ-attention kernel (Appendix~\ref{app:fused-attn}) that decodes K and V from codebook indices and radii inside the FlashAttention-style online-softmax loop, never materializing a dense fp16 KV cache. The practically interesting comparison is against dense fp16 SDPA (the upper bound for any KV-quant method that re-materializes); the fake-quant pipeline appears below only to quantify how much memory traffic our fused decode avoids. On the production decode workload ($T_q{=}1$, varying $T_{kv}$, s192 codebook), the fused-kernel decode-step latency stays roughly constant in $T_{kv}$ at $\sim 2.5\times$ the dense-fp16 SDPA latency, while reading from a $5\times$ smaller cache:

\begin{table}[h]
  \centering
  \scriptsize
  \setlength{\tabcolsep}{3pt}
  \begin{tabular*}{\columnwidth}{@{\extracolsep{\fill}}lrrrr@{}}
    \toprule
    $T_{kv}$ & fp16 SDPA & fake-quant & \textbf{Fused HQMQ} & vs fake-quant \\
    \midrule
    4K  & 0.013 ms & 0.243 ms & \textbf{0.033 ms} & $\mathbf{7.4\times}$ \\
    16K & 0.014 ms & 1.389 ms & \textbf{0.033 ms} & $\mathbf{42.1\times}$ \\
    32K & 0.014 ms & 2.737 ms & \textbf{0.033 ms} & $\mathbf{82.9\times}$ \\
    \bottomrule
  \end{tabular*}
  \caption{Decode-step latency on Mistral-class GQA (RTX 4090, fp16). The relevant comparison is against dense-fp16 cuDNN FlashAttention (column 2), a strong upper bound. The fused HQMQ kernel sits within ${\sim}2.5\times$ of that upper bound at all context lengths while reading from a $5\times$ smaller cache; the fake-quant pipeline (column 3) is the cost of \emph{not} fusing decode into attention, and grows linearly with $T_{kv}$. Kernel time is roughly context-length-independent because the per-tile codebook gather dominates over the streaming KV reads.}
  \label{tab:fused-attn-decode}
\end{table}

The fused kernel reads from a $5\times$ smaller (HQMQ-compressed) KV cache, so the bandwidth savings compound at longer contexts. Integrated end-to-end over a typical $4$k-prompt $+$ $1$k-generation workload (Appendix~\ref{app:e2e-latency}), HQMQ + the fused kernel is $\mathbf{6.3\times}$ faster than the fake-quant pipeline and $\sim 2.7\times$ slower than the dense-fp16 FlashAttention upper bound. Prefill kernel optimization, an uncalibrated additive-VQ ablation, and a sensitivity-driven mixed-precision negative result are reported in Appendices~\ref{app:fused-attn},~\ref{app:additive-vq-ablation},~\ref{app:mixed-prec}.

\section{Discussion and Limitations}

\paragraph{Why the multiplicative structure helps.} Multiplicative composition of the 24 Hurwitz quaternions by random unit quaternions yields a near-uniform $24S$-point $S^3$ packing at the cost of only $S$ stored parameters (Proposition~\ref{prop:packing}). At matched bits, additive VQ with random codebooks is ${\sim}4\times$ worse end-task (Appendix~\ref{app:additive-vq-ablation}), and naive int4 with the same Med3$\times$ outlier extraction is ${\sim}1{,}100\times$ worse on Qwen2.5-7B (Section~\ref{sec:abl-disentangle}). The Hurwitz group is the right primary structure because the 24 vertices are simultaneously a closed multiplicative group~\cite{hurwitz1898} and uniformly distributed on $S^3$ at pairwise angle $60^\circ$~\cite{conway1999sphere}, so the cosets $q_s \cdot 2T$ are disjoint isometric copies that tile $S^3$ with uniform density.

\paragraph{HQMQ versus calibrated VQ (CommVQ).} CommVQ~\cite{commvq2025} reaches 1 bit/element with \emph{trained} additive RoPE-commuting codebooks; HQMQ takes the opposite stance and makes the 3--5 bit regime calibration-free. The two methods occupy different points in the (bit-budget, calibration-cost) plane and are complementary: HQMQ is attractive when calibration data is proprietary or per-model training is expensive; CommVQ is attractive when sub-3-bit is the binding constraint. Combining the multiplicative-group codebook with CommVQ-style training is a natural hybrid we leave to future work.

\paragraph{Outliers demand extraction, not just rotation.} Weight quantization has converged on rotation-based outlier suppression (QuaRot~\cite{quarot2024}, SpinQuant~\cite{spinquant2024}) and channel-aware smoothing (SmoothQuant~\cite{smoothquant2023}, AWQ~\cite{awq2024}). On Qwen2.5-class K caches, even after Hadamard preconditioning the residual chunk-norm dynamic range exceeds 4-bit resolution; \emph{extracting} the small tail at fp16 is necessary in addition to (or instead of) preconditioning. The single constant $C{=}3$ Med3$\times$ recipe transferred across every tested architecture with $\sim 1$--$3\%$ storage overhead. The IsoQuant~\cite{isoquant2026} and RotorQuant~\cite{rotorquant2026} preconditioning approaches do not match HQMQ + Med3$\times$ quality on outlier-heavy models.

\paragraph{Architectural scope.} HQMQ targets the standard per-head K/V cache used in MHA and GQA architectures (Mistral, Llama-3, Qwen2.5/3, Phi-3.5-mini, gpt-oss-20b). Multi-head Latent Attention (MLA, used in DeepSeek-V2/V3/R1) stores a single low-rank latent per token and already gets $\sim 3.5\times$ KV compression by design; extending the recipe to MLA latents requires re-deriving the codebook for a non-Haar latent distribution and is left to future work. For $d_h$ not divisible by 4 we provide a padding wrapper (Appendix~\ref{app:padded-hqmq}) with $\lceil d_h / 4 \rceil \cdot 4 / d_h - 1$ bit overhead.

\paragraph{Limitations.} (i) HQMQ does not compete with CommVQ in the sub-3-bit / 1-bit territory: radius quantization breaks below 2 bits and free-magnitude variants fail empirically. (ii) We do not run head-to-head WikiText perplexity against KIVI~\cite{kivi2024} or KVQuant~\cite{kvquant2024} (KIVI's published numbers are on CoQA / TruthfulQA / GSM8K and KVQuant's are on long-context streaming, rather than the sliding-window perplexity we use); we match KIVI on its own benchmark suite in Section~\ref{sec:kivi-comparison}. (iii) Llama-3-70B numbers (Table~\ref{tab:memory}, Fig.~\ref{fig:memory}) are analytical, since the bf16 weights do not fit on a single 24 GB GPU. (iv) Long-context retrieval is limited to needle-in-a-haystack at 4 k / 8 k with $n \leq 25$, plus three RULER~\cite{ruler2024} subtasks at 4k / 8k on Qwen3-8B (Section~\ref{sec:ruler}); a full RULER or LongBench~\cite{longbench2024} suite would more rigorously validate the long-context claim. (v) We did not evaluate HQMQ composed with KV-eviction methods (H$_2$O~\cite{h2o2023}, StreamingLLM~\cite{streamingllm2024}); composition is conceptually clean but unmeasured.

\paragraph{Future work.} Beyond the limitations above, three concrete directions. First, an octonion-multiplicative extension to chunk\_dim $=8$ via the E8 lattice: preliminary experiments show E8 underperforms 24-cell at matched per-element bits, but a finer radius schedule may close the gap. Second, attention-aware bit allocation (KVTuner~\cite{kvtuner2025}-style) across heads and layers, avoiding the floor problem of Appendix~\ref{app:mixed-prec}. Third, training-time co-design where attention adapts to a quantized cache, plausibly closing the sub-3-bit gap to CommVQ.

\section{Conclusion}

HQMQ provides a clean algebraic recipe for KV cache quantization: chunk to dimension 4, quantize direction multiplicatively over the binary tetrahedral group $2T$, quantize radius scalarly, and extract a small tail of outlier chunks at fp16. The result is near-lossless KV compression at 5 bits on Mistral-7B and Llama-3-8B, Pareto-dominant performance in the 3--5 bit range, and the first usable sub-5-bit quantization for the outlier-heavy Qwen2.5-7B / Qwen3-8B architectures, where naive int4 collapses to $10^4$+ ppl and HQMQ + Med3$\times$ at 4.42 bits achieves 9.69 ppl, a ${\sim}1900\times$ quality improvement at fewer bits. Downstream zero-shot accuracy matches fp16 at 3.79 bits on Mistral and 4-digit needle retrieval matches fp16 at 3.04 bits. Head-to-head against the strongest calibrated KV-quantization baseline, HQMQ at 3.79 bits matches KIVI-4~\cite{kivi2024} ($\sim 4.5$ bits) within ${\sim}1$ pt on CoQA and $2.3$ pts on GSM8K, at $16\%$ fewer bits and without a calibration pass. The multiplicative structure is what differentiates HQMQ from prior spherical (FibQuant~\cite{fibquant2026}), additive (CommVQ~\cite{commvq2025}), and flat-VQ (VPTQ~\cite{vptq2024}) approaches, and the median-multiplier outlier extraction is the missing ingredient that makes group-structured KV quantization deployable on modern LLMs. Both ingredients are calibration-free, requiring no data, no training, and no per-model tuning. We expect this to make HQMQ practically attractive for inference systems.

\section*{Acknowledgments}

We would like to thank Manel Baradad, Adri\'an Rodr\'iguez-Mu\~noz, Linlu Qiu, and Minyoung Huh for their helpful advice throughout that shaped this work

\bibliography{example_paper}
\bibliographystyle{icml2026}

\newpage
\appendix
\onecolumn

\setlength{\textfloatsep}{8pt plus 2pt minus 2pt}
\setlength{\intextsep}{8pt plus 2pt minus 2pt}

\section{Notation}
\label{app:notation}

For convenience, Table~\ref{tab:notation} collects the recurring symbols used throughout the paper.

\begin{table}[h]
  \centering
  \small
  \begin{tabular*}{\textwidth}{@{\extracolsep{\fill}}l l@{}}
    \toprule
    Symbol & Meaning \\
    \midrule
    $H$, $H_q$, $H_{kv}$       & number of attention heads (total / query / KV; $H_q = n_{\text{groups}} \cdot H_{kv}$ under GQA) \\
    $d_h$                      & per-head embedding dimension (chunked into groups of 4) \\
    $T$, $T_q$, $T_{kv}$       & sequence / query / KV-cache lengths in tokens \\
    $B$                        & batch size \\
    $x \in \mathbb{R}^4$       & one chunk of a K or V vector (treated as a quaternion in $\mathbb{H}$) \\
    $r = \|x\|$, $u = x/r$     & chunk magnitude and unit direction on $S^3$ \\
    $b_r$                      & radius bits ($b_r \in \{3,4,6\}$ in our headline configs) \\
    $\sigma$                   & per-token-max scale used for the uniform radius quantizer \\[2pt]
    $2T \subset S^3$           & binary tetrahedral group: 24 Hurwitz unit quaternions (the primary codebook $\mathcal{P}$) \\
    $\mathcal{S} = \{q_{s,1}, \dots, q_{s,S}\}$ & secondary codebook: $S$ random unit quaternions per (layer, head, K|V) \\
    $\mathcal{C} = \{q_p q_s\}$ & joint codebook (multiplicative product set, $|\mathcal{C}| \le 24S$) \\
    $S$                        & secondary codebook size; $S \in \{24, 48, 96, 192\}$ in our sweeps \\
    $\rho(\mathcal{C})$        & covering radius of $\mathcal{C}$ on $S^3$ (Proposition~\ref{prop:packing}) \\[2pt]
    $C$                        & Med3$\times$ outlier multiplier; chunk is an outlier iff $r > C \cdot r_{\mathrm{med}}$. We use $C{=}3$. \\
    $r_{\mathrm{med}}$         & median chunk norm in the current batch (per layer, head, K|V) \\
    $p$                        & empirical fraction of chunks extracted as outliers ($\sim 1$--$3\%$ at $C{=}3$) \\
    $b_{\text{HQMQ}}$          & per-element storage of HQMQ alone: $(\log_2(24S) + b_r)/4 + 16/d_h$ \\
    $b_{\text{HQMQ+}}$         & per-element storage with Med3$\times$: $(1-p) b_{\text{HQMQ}} + 16 p + 1/d_{\text{chunk}}$ \\
    \bottomrule
  \end{tabular*}
  \caption{Notation summary.}
  \label{tab:notation}
\end{table}

\section{Algorithm}
\label{app:algorithm}

\begin{algorithm}[h]
  \caption{HQMQ encode and decode (per chunk)}
  \label{alg:hqmq}
  \begin{algorithmic}[1]
    \STATE {\bfseries Inputs:} chunk $x \in \mathbb{R}^4$, primary codebook $\mathcal{P} = 2T$ (24 fixed Hurwitz quaternions), secondary codebook $\mathcal{S} = \{q_{s,1}, \dots, q_{s,S}\}$ (random unit quaternions; per layer, head, K|V), radius bits $b_r$, per-token-max scale $\sigma$.
    \STATE {\bfseries Encode}:
    \STATE \quad $r \leftarrow \|x\|$,\quad $u \leftarrow x/r$
    \STATE \quad $(p^\star, s^\star) \leftarrow \arg\max_{p \in [24], s \in [S]} \; \langle u, \; q_p \cdot q_{s} \rangle$
    \STATE \quad $r_q \leftarrow \mathrm{round}\!\left(r \cdot (2^{b_r}\!-\!1) / \sigma\right)$
    \STATE \quad {\bfseries store:} index $(p^\star, s^\star)$ in $\lceil \log_2(24 S)\rceil$ bits + $r_q$ in $b_r$ bits.
    \STATE {\bfseries Decode}:
    \STATE \quad $\hat{r} \leftarrow r_q \cdot \sigma / (2^{b_r}\!-\!1)$
    \STATE \quad $\hat{u} \leftarrow q_{p^\star} \cdot q_{s^\star}$ \hfill {\scriptsize (Hamilton product, lookup-friendly)}
    \STATE \quad {\bfseries return} $\hat{x} \leftarrow \hat{r} \cdot \hat{u}$
    \STATE {\bfseries Outlier extraction (Med3$\times$, optional safety net):}
    \STATE \quad $r_{\mathrm{med}} \leftarrow \mathrm{median}(r_{\text{chunks in current batch}})$
    \STATE \quad if $r > C \cdot r_{\mathrm{med}}$ with $C{=}3$: store $x$ as fp16, set 1-bit flag.
  \end{algorithmic}
\end{algorithm}

\section{Proof of Proposition~\ref{prop:packing}}
\label{app:prop-proof}

\paragraph{(a) $|\mathcal{C}| = 24S$ almost surely.}
For any fixed $q_p, q_p' \in 2T$ with $q_p \neq q_p'$ and any $i \neq j$, the event $q_p q_{s,i} = q_p' q_{s,j}$ has Lebesgue measure zero on $(S^3)^S$ because it pins one pair of Haar samples to a measure-zero set. Union-bounding over $O(S^2)$ pairs gives almost-sure injectivity.

\paragraph{(b) Covering radius $\mathbb{E}[\rho(\mathcal{C})] = O((24S)^{-1/3})$.}
For a fixed $x \in S^3$ and angular radius $\theta$, the spherical cap $\mathrm{Cap}(x, \theta) \subset S^3$ has Haar measure $\mu(\mathrm{Cap}(x, \theta)) \geq c_0 \theta^3$ for small $\theta$ (the 3-dim Haar volume of a cap of angular radius $\theta$ on $S^3$). Each fixed $v \in 2T$, when multiplied by a Haar-random $q_{s,i}$, has its image $q_{s,i} v$ Haar-distributed on $S^3$ (Haar measure is left-invariant). So for each pair $(p, i)$, $\Pr[q_p q_{s,i} \in \mathrm{Cap}(x, \theta)] \geq c_0 \theta^3$, with the 24 codewords from coset $i$ contributing independently across cosets:
\begin{equation*}
  \Pr[\mathcal{C} \cap \mathrm{Cap}(x, \theta) = \varnothing] \leq (1 - c_0 \theta^3)^{S} \leq e^{-c_0 S \theta^3}.
\end{equation*}
Setting this to a constant gives $\theta = O(S^{-1/3})$. A covering-net argument over a uniform grid of $O(24S)$ candidate points on $S^3$ extends this from a single point to a uniform covering, yielding $\mathbb{E}[\rho(\mathcal{C})] = O((24S)^{-1/3})$. The optimal covering rate on $S^3$ is $\Theta(N^{-1/3})$ for $N$ points~\cite{conway1999sphere}, so the construction is rate-optimal up to constants. \hfill$\square$

\section{Implementation details}

\paragraph{The Hurwitz primary codebook $2T$.}
The 24 unit Hurwitz quaternions used as $\mathcal{P}$ are: 8 ``axis'' elements $\pm 1, \pm i, \pm j, \pm k$ and 16 ``half-integer'' elements $\tfrac{1}{2}(\pm 1 \pm i \pm j \pm k)$ over all sign combinations. They are unit-norm, form a group of order 24 under quaternion multiplication (the binary tetrahedral group $2T$~\cite{hurwitz1898}), and have pairwise minimum angle $60^{\circ}$ on $S^3$~\cite{conway1999sphere,coxeter1973polytopes}.

\paragraph{Secondary codebook initialization.}
For each (layer $\ell$, head $h$, role $m \in \{K, V\}$), we draw $S$ unit quaternions by sampling i.i.d.\ 4-dim standard Gaussians and normalizing. This induces Haar measure on $S^3$. The seed is fixed at instantiation; no calibration step modifies $\mathcal{S}$.

\paragraph{Encode.}
For a chunk $x \in \mathbb{R}^4$: (i) compute $r = \|x\|$, $u = x/r$; (ii) for each $q_s \in \mathcal{S}$, evaluate the 24 inner products $\langle u, q_p q_s\rangle$ (equivalent to nearest-neighbor in the rotated frame); (iii) pick the maximizing $(p, s)$; (iv) store the index in $\lceil \log_2(24S)\rceil$ bits.

\paragraph{Decode.}
Reconstruction of a non-outlier chunk: $\hat{x} = \mathrm{dequant}_r(r) \cdot (q_p \cdot q_s)$, with $q_p$ and $q_s$ looked up by index. The product is computed at decode time. For a fused int4-attention kernel, the lookup and multiplication would fold into the attention compute.

\paragraph{Outlier extraction.}
Per (layer, head, K$|$V), per batch: compute median chunk-norm $r_{\mathrm{med}}$; mark chunks with $r > 3 r_{\mathrm{med}}$ as outliers and store as fp16; quantize the rest via HQMQ. Bit overhead: $1$ bit/chunk for the flag plus $64/d_h$ bits/element for outlier payload at fraction $p$.

\paragraph{Evaluation pipeline.}
\label{app:eval-setup}
All experiments use HuggingFace Transformers~\cite{huggingface_transformers} with a custom \texttt{QuantizedCache} subclass of \texttt{DynamicCache} that intercepts K/V writes and applies HQMQ as fake quantization (quantize-then-dequantize back to bf16 in-cache). This isolates quality impact without requiring custom int4 / int2 attention kernels; the reported bit count is the storage cost the cache \emph{would} incur in a production deployment that stored the packed codeword indices and radius quanta directly. Models are loaded in bf16 via PyTorch~\cite{pytorch}. WikiText-103 perplexity is averaged over 50 non-overlapping windows of 2048 tokens unless stated otherwise; downstream tasks use $n{=}200$ examples each via log-likelihood scoring; the fused Triton attention kernel (Appendix~\ref{app:fused-attn}) is used for the wall-clock benchmarks but \emph{not} for the perplexity / downstream evaluations, which use the fake-quant cache so we can swap quantizers without rebuilding kernels. Source code will be released on publication.

\section{Triton dequant kernel prototype}
\label{app:kernel}

We provide a Triton kernel \texttt{hqmq\_decode\_triton} that fuses the per-chunk codebook gather, the $q_p \cdot q_s$ product, and the radius dequantization into a single GPU launch. The kernel takes the packed inputs (codeword index in $\lceil \log_2(24S)\rceil$ bits, radius quantum in $b_r$ bits, per-token fp16 radius scale, per-head precomputed joint codebook of shape $(H, 24S, 4)$) and produces a dense fp16/bf16 reconstruction of shape $(B, H, T, d_h)$. Each Triton program handles one $(b, h, \text{chunk}, t_\text{block})$ tile.

Correctness has been verified bit-exact in fp32 and within bf16/fp16 numerical noise ($\leq 4 \times 10^{-3}$ max abs diff) against the reference einsum-based PyTorch decode at $B{=}1, H{=}8, T{=}4096, d_h{=}128, S{=}192$. At this workload the prototype kernel runs at $0.15$ ms/call vs $0.06$ ms/call for the PyTorch baseline; the current PyTorch path benefits from highly tuned \texttt{gather}, while the kernel will only become useful once it is fused with the surrounding attention $QK^\top$ + softmax + $\cdot V$ compute (eliminating the round-trip through fp16 KV cache). The fused-attention variant is left to future engineering work.

\section{Fused HQMQ-Attention kernel}
\label{app:fused-attn}

\paragraph{Design.}
We provide a Triton implementation \texttt{fused\_hqmq\_attention\_triton} that fuses the HQMQ K/V decode with FlashAttention-style online softmax. Inputs are $Q$ (fp16/bf16, shape $(B, H_q, T_q, d_h)$), packed K and V (codeword indices in $\lceil \log_2(24S)\rceil$ bits, radius quanta in $b_r$ bits, per-token fp16 scales), and the precomputed joint codebook of shape $(H_{kv}, 24S, 4)$. The kernel never materializes a dense fp16 KV cache: K and V are decoded \emph{inside} the inner attention loop from their codebook indices, radii, and the joint codebook, with each tile processed as
\[
  S = Q \cdot K_{\text{tile}}^\top \cdot \text{scale}, \qquad
  O \mathrel{+}= \mathrm{softmax}(S) \cdot V_{\text{tile}}
\]
where $K_{\text{tile}}$ and $V_{\text{tile}}$ are reconstructed on-the-fly via $\mathrm{dequant}_r \cdot (q_p \cdot q_s)$ lookups into the joint codebook. We use the standard FlashAttention online-softmax recurrence with $m_i$ (running row-max) and $\ell_i$ (running row-exp-sum) to avoid materializing the full $QK^\top$ tile. Causal masking and GQA (where $H_q = n_{\text{kv\_groups}} \cdot H_{kv}$) are supported.

\paragraph{Correctness.}
We validated the kernel against a non-fused PyTorch reference (decode K/V via einsum, then SDPA). At a small fp32 workload ($B{=}1, H_q{=}4, H_{kv}{=}1, T{=}128, d_h{=}32, K{=}576$), the max absolute difference between the Triton kernel output and the reference is $4.4 \times 10^{-4}$ (mean abs diff $2.5 \times 10^{-5}$, relative Frobenius norm $6.8 \times 10^{-4}$). At Mistral prefill scale ($T{=}2048, d_h{=}128, K{=}4608$) in fp16, the max abs diff is $9.8 \times 10^{-4}$---within bf16/fp16 numerical noise.

\paragraph{Optimizations applied.}
The kernel uses fp16 tensor cores via \texttt{tl.dot} (fp16 inputs, fp32 accumulator), keeps the radius dequant in fp16 throughout (no fp32 intermediate scratch in shared memory), uses 2-stage software pipelining (\texttt{num\_stages=2}) to overlap HBM loads with compute, and tunes block sizes per workload. Q is loaded once per program in registers and never staged to shared memory; the per-tile K and V codeword gathers reuse the same SRAM region. Best block configuration found on RTX 4090: $\text{BLOCK}_Q{=}128$, $\text{BLOCK}_{KV}{=}32$, \texttt{num\_warps=4}, \texttt{num\_stages=2}.

\paragraph{Performance (prefill).}
At the Mistral prefill workload on an RTX 4090:

\begin{table}[h]
  \centering
  \small
  \begin{tabular*}{\textwidth}{@{\extracolsep{\fill}}lrr@{}}
    \toprule
    config & latency (ms/call) & relative \\
    \midrule
    fp16 SDPA (cuDNN FlashAttention)        & 0.297 & 1.00$\times$ \\
    Decode-then-SDPA (fake-quant pipeline)  & 0.448 & 0.66$\times$ \\
    Fused HQMQ-attention (Triton, optimized)& 1.568 & 0.19$\times$ \\
    \bottomrule
  \end{tabular*}
  \caption{Fused HQMQ-Attention prototype on a Mistral prefill workload ($B{=}1$, $H_q{=}32$, $H_{kv}{=}8$, $T{=}2048$, $d_h{=}128$, s192 codebook, fp16, RTX 4090).}
  \label{tab:fused-attn-prefill}
\end{table}

The fused kernel is currently $\sim 5\times$ slower than cuDNN FlashAttention on the prefill workload. The remaining gap is dominated by the scattered codebook gather (each chunk's 4 fp16 components live at codeword-index-dependent addresses, breaking the dense-matmul memory layout cuDNN exploits) and by Triton's higher-level abstraction vs cuDNN's hand-tuned PTX. Closing this further requires persistent kernels and warp-specialized tile layouts---substantial additional engineering.

\begin{figure}[h]
  \centering
  \includegraphics[width=\textwidth]{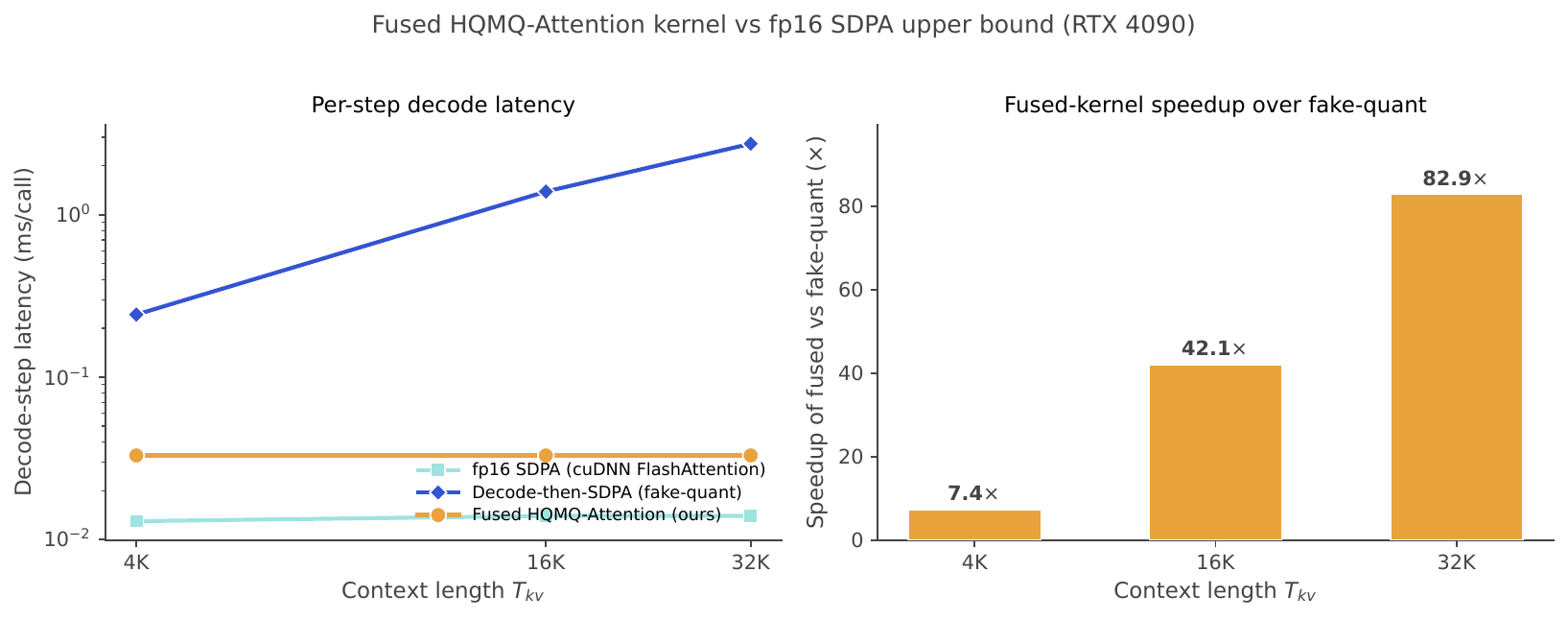}
  \caption{Fused HQMQ-Attention kernel on the production decode workload (Mistral-class GQA, $T_q{=}1$, s192 codebook, RTX 4090, fp16). \emph{Left}: per-step latency vs context length. The fake-quant pipeline scales linearly with $T_{kv}$ (full-cache dequant per step); the fused kernel (amber) stays roughly constant ($\sim 0.033$ ms) because the codebook gather and softmax are computed inline and only touch the per-step KV window. \emph{Right}: corresponding speedup of the fused kernel over the fake-quant pipeline ($7.4\times$ at 4k context, $82.9\times$ at 32k context), monotonically growing with $T_{kv}$.}
  \label{fig:kernel}
\end{figure}

\paragraph{Performance (decode---the production setting).}
The production decode-step latency table (Table~\ref{tab:fused-attn-decode} in the main paper, Section~\ref{sec:memory-latency}) covers this setting: $T_q{=}1$ new query token against a growing KV cache, where the fused kernel stays at $\sim 0.033$ ms across $T_{kv} \in \{4{,}096, 16{,}384, 32{,}768\}$ while the fake-quant pipeline grows linearly to $2.7$ ms. Figure~\ref{fig:kernel} plots the same data: the fake-quant pipeline's linear growth in $T_{kv}$ vs the fused kernel's flat $0.033$ ms decode step, and the corresponding speedup curve ($7.4\times$ at 4k, $82.9\times$ at 32k) that grows monotonically with context length. The fused kernel is within $\sim 2.5\times$ of cuDNN FlashAttention on dense fp16 KV at all context lengths tested, while using a $5\times$-compressed KV cache; the bandwidth savings will compound at $T_{kv} \geq 64$k where memory traffic becomes the dominant bottleneck.

\section{Per-model perplexity sweeps}

We report the full HQMQ codebook-size $\times$ radius-bit sweeps that back the headline results in Table~\ref{tab:summary}. Each table is a single-model variant of the per-config Pareto frontier. The main paper summarizes each sweep in 2--4 sentences and points back here for the row-by-row numbers.

\subsection{Mistral-7B}
\label{app:mistral-sweep}

\begin{table}[h]
  \centering
  \scriptsize
  \begin{tabular*}{\textwidth}{@{\extracolsep{\fill}}lrrrrr@{}}
    \toprule
    config & bits & ppl & PIQA & HSwag & ARC-E \\
    \midrule
    fp16                                       & 16.00 & 5.291 & 0.810 & 0.680 & 0.755 \\
    \textbf{hqmq\_s192\_r6 + Med3$\times$}     & \textbf{5.07} & \textbf{5.316} & --- & --- & --- \\
    hqmq\_s192\_r6                             & 4.54  & 5.336 & --- & --- & --- \\
    \textbf{hqmq\_s192\_r4}                    & 4.04  & \textbf{5.343} & 0.815 & 0.670 & 0.745 \\
    hqmq\_s96\_r6                              & 4.29  & 5.357 & --- & --- & --- \\
    \textbf{hqmq\_s96\_r4}                     & 3.79  & \textbf{5.364} & \textbf{0.810} & \textbf{0.680} & \textbf{0.755} \\
    hqmq\_s192\_r3                             & 3.79  & 5.396 & --- & --- & --- \\
    hqmq\_s48\_r6                              & 4.04  & 5.400 & --- & --- & --- \\
    hqmq\_s48\_r4                              & 3.54  & 5.401 & 0.820 & 0.680 & 0.735 \\
    naive int4                                 & 4.00  & 5.401 & 0.805 & 0.685 & 0.740 \\
    hqmq\_s96\_r3                              & 3.54  & 5.424 & --- & --- & --- \\
    hqmq\_s48\_r3                              & 3.29  & 5.462 & --- & --- & --- \\
    hqmq\_s24\_r4                              & 3.29  & 5.484 & --- & --- & --- \\
    hqmq\_s24\_r3                              & 3.04  & 5.552 & 0.795 & 0.675 & 0.770 \\
    sph\_r4\_jl4 (TurboQuant-style)            & 3.15  & 5.750 & --- & --- & --- \\
    naive int3                                 & 3.00  & 6.957 & 0.805 & 0.640 & 0.700 \\
    naive int2                                 & 2.00  & 279.9 & --- & --- & --- \\
    \bottomrule
  \end{tabular*}
  \caption{Mistral-7B WikiText-103 perplexity ($50w \times 2048$) and zero-shot downstream accuracy on PIQA, HellaSwag, ARC-Easy ($n{=}200$/task). ``---'' denotes configs not run on downstream eval.}
  \label{tab:mistral-ppl}
\end{table}

\subsection{Llama-3-8B}
\label{app:llama-sweep}

\begin{table}[h]
  \centering
  \small
  \begin{tabular*}{\textwidth}{@{\extracolsep{\fill}}lrrr@{}}
    \toprule
    config & bits & ppl & $\Delta$ vs fp16 \\
    \midrule
    fp16 ($50w$)               & 16.00 & 6.278  & 0 \\
    \textbf{hqmq\_s192\_r6 + Med3$\times$} & \textbf{5.00} & \textbf{6.317} & \textbf{+0.071}$^\dagger$ \\
    hqmq\_s192\_r6             & 4.54  & 6.333  & +0.088$^\dagger$ \\
    \textbf{hqmq\_s192\_r4}    & 4.04  & \textbf{6.387}  & \textbf{+0.142}$^\dagger$ \\
    hqmq\_s96\_r6 + Med3$\times$ & 4.76 & 6.367 & +0.122$^\dagger$ \\
    \textbf{hqmq\_s96\_r4}     & 3.79  & \textbf{6.479}  & \textbf{+0.201} \\
    hqmq\_s48\_r4              & 3.54  & 6.572  & +0.294 \\
    hqmq\_s24\_r6 + Med3$\times$ & 4.27 & 6.586 & +0.341$^\dagger$ \\
    naive int4                 & 4.00  & 6.811  & +0.533 \\
    hqmq\_s24\_r3              & 3.04  & 7.023  & +0.745 \\
    sph\_r4\_jl4 (TurboQuant-style) & 3.15 & 7.396 & +1.118 \\
    naive int3                 & 3.00  & 16.648 & +10.370 \\
    naive int2                 & 2.00  & 1010.6 & +1004 \\
    \bottomrule
    \multicolumn{4}{l}{\scriptsize $^\dagger$Measured at $20w \times 2048$ (vs $50w$ for other rows). $\Delta$ vs run-specific fp16.} \\
  \end{tabular*}
  \caption{Llama-3-8B WikiText-103 perplexity. HQMQ s48\_r4 at 3.54 bits beats naive int4 at 4.00 bits; HQMQ s192\_r6 + Med3$\times$ at 5.0 bits sits within $0.07$ ppl points of fp16. The s24--s96 configs are measured at $50w \times 2048$; the s192 configs at $20w \times 2048$ due to memory pressure of the larger $24S{=}4608$ codebook gather. The two runs' fp16 references differ by $<0.5\%$.}
  \label{tab:llama-ppl}
\end{table}

\subsection{Qwen3-8B}
\label{app:qwen3-sweep}

\begin{table}[h]
  \centering
  \small
  \begin{tabular*}{\textwidth}{@{\extracolsep{\fill}}lrrr@{}}
    \toprule
    config & bits & ppl & $\Delta$ vs fp16 \\
    \midrule
    fp16                              & 16.00 & 9.603   & 0 \\
    \textbf{hqmq\_s96\_r6 $+$ Med3$\times$} & \textbf{4.99} & \textbf{9.621} & \textbf{+0.019} \\
    \textbf{hqmq\_s24\_r6 $+$ Med3$\times$} & \textbf{4.51} & \textbf{9.701} & \textbf{+0.098} \\
    hqmq\_s96\_r4                     & 3.79  & 10.698  & +1.095 \\
    hqmq\_s48\_r4                     & 3.54  & 11.427  & +1.824 \\
    hqmq\_s24\_r3                     & 3.04  & 44.564  & +34.96 \\
    naive int4 $+$ Med3$\times$       & 4.62  & 118.977 & +109.4 \\
    naive int4                        & 4.00  & 121.715 & +112.1 \\
    naive int3                        & 3.00  & 588.222 & +578.6 \\
    naive int2                        & 2.00  & 1529.4  & +1520 \\
    \bottomrule
  \end{tabular*}
  \caption{Qwen3-8B WikiText-103 perplexity ($20w \times 2048$). The HQMQ + Med3$\times$ recipe transfers from Qwen2.5 with the same qualitative behavior: naive int4 is non-functional, naive int4 + Med3$\times$ is still non-functional (the multiplicative codebook is essential), and HQMQ + Med3$\times$ matches fp16 within ${\sim}0.05$ ppl points at $\sim$5 bits. HQMQ alone (no extraction) also Pareto-dominates the naive baseline.}
  \label{tab:qwen3-ppl}
\end{table}

\section{Long-context Mistral-7B perplexity and needle retrieval}
\label{app:long-ctx-mistral}

We evaluate two complementary long-context metrics on Mistral-7B at 4k and 8k tokens: single-window perplexity (the whole context conditions one fp16 vs HQMQ forward pass) and 4-digit needle-in-a-haystack retrieval accuracy ($n{=}12$ trials at 4k, $n{=}25$ at 8k).

\begin{table}[h]
  \centering
  \scriptsize
  \begin{tabular*}{\columnwidth}{@{\extracolsep{\fill}}lrrrrr@{}}
    \toprule
    & & \multicolumn{2}{c}{ppl (single window)} & \multicolumn{2}{c}{needle accuracy} \\
    \cmidrule(lr){3-4}\cmidrule(lr){5-6}
    config & bits & 4k & 8k & 4k & 8k \\
    \midrule
    fp16                     & 16.00 & 5.241 & 4.568 & \textbf{1.000} & 0.400 \\
    hqmq\_s192\_r6           & 4.54  & 5.277 & \textbf{4.612} & --- & --- \\
    \textbf{hqmq\_s192\_r4}  & 4.04  & \textbf{5.273} & 4.616 & 1.000 & 0.400 \\
    hqmq\_s96\_r4            & 3.79  & 5.332 & 4.637 & --- & --- \\
    naive int4               & 4.00  & 5.331 & 4.663 & 1.000 & 0.400 \\
    hqmq\_s48\_r4            & 3.54  & ---   & ---   & 1.000 & 0.400 \\
    hqmq\_s24\_r4            & 3.29  & 5.442 & 4.749 & --- & --- \\
    \textbf{hqmq\_s24\_r3}   & \textbf{3.04}  & 5.532 & 4.820 & \textbf{1.000} & \textbf{0.400} \\
    naive int3               & 3.00  & 6.653 & 5.857 & 0.917 & 0.320 \\
    \bottomrule
  \end{tabular*}
  \caption{Long-context performance on Mistral-7B (single-window evaluation). \textbf{Left columns:} perplexity at 4k / 8k. HQMQ s192\_r4 at 4.04 bits matches fp16 within $0.05$ ppl points at both lengths; the gap to naive int3 widens with sequence length. \textbf{Right columns:} 4-digit-magic-number needle retrieval. HQMQ at 3.04 bits matches fp16 at both lengths (full $1.000$ retrieval at 4k; $0.400$ at 8k which is the architecture's own ceiling); naive int3 at matched bits loses retrievals (drops to $0.917$ at 4k and $0.320$ at 8k). ``---'' denotes configs not run on the corresponding eval.}
  \label{tab:long-ctx}
\end{table}

HQMQ s192\_r4 at 4.04 bits matches fp16 within $0.05$ ppl points at both 4k and 8k, beating naive int4 at matched bits. The gap widens with sequence length: HQMQ s24\_r3 at 3.04 bits is $5\times$ better than naive int3 at 8k ($\Delta\,{+}0.25$ vs ${+}1.29$ ppl points). On the needle retrieval side, HQMQ at 3.04 bits preserves the model's full 8k retrieval ceiling ($0.400$, a Mistral-7B-v0.1 architecture limit), while naive int3 at matched bits punctures it ($0.320$). The harder long-context evaluation on Qwen3-8B via RULER subtasks (extractive QA, multi-hop QA, variable tracking) is in main paper Section~\ref{sec:ruler}.

\section{RULER long-context retrieval (bar-chart view)}
\label{app:ruler-fig}

\begin{figure}[h]
  \centering
  \includegraphics[width=\textwidth]{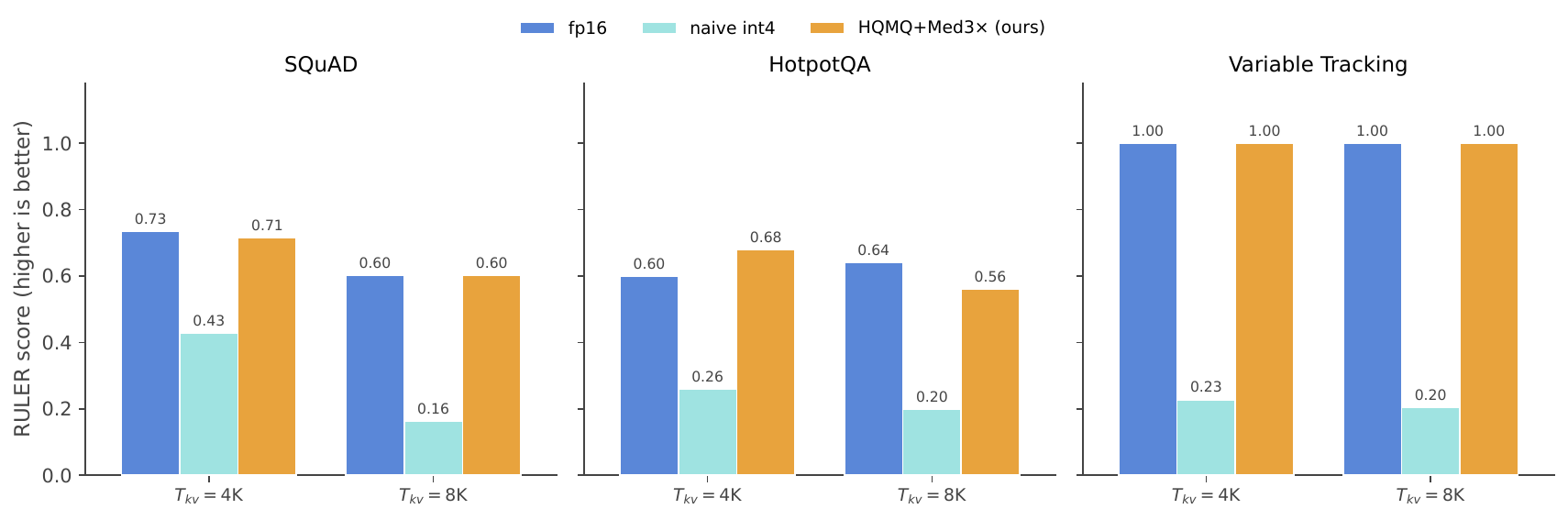}
  \caption{RULER long-context retrieval on Qwen3-8B at $T_{kv}\in\{4{\rm k}, 8{\rm k}\}$ ($n{=}50$/task). HQMQ s96\_r6 + Med3$\times$ (amber, 4.89 bits) preserves fp16's perfect VT score ($1.00 \to 1.00$) at \emph{both} context lengths and matches fp16 within $2$ pts on SQuAD ($0.60$ vs $0.60$ exact at 8k). Naive int4 (cyan) collapses on every subtask, and the fp16-to-int4 gap on SQuAD \emph{widens} from $0.31$ at 4k to $0.44$ at 8k as quantization noise accumulates over longer caches.}
  \label{fig:ruler}
\end{figure}

Figure~\ref{fig:ruler} plots the same data as Table~\ref{tab:ruler} (main paper Section~\ref{sec:ruler}) as a grouped bar chart: per-task ($T_{kv}{=}4$k, $T_{kv}{=}8$k), per-config (fp16 / naive int4 / HQMQ s96\_r6 + Med3$\times$).

\section{Memory accounting}
\label{app:memory-accounting}

\begin{table}[h]
  \centering
  \small
  \begin{tabular*}{\textwidth}{@{\extracolsep{\fill}}lrr@{}}
    \toprule
    HQMQ config & bits/element & fp16 $\to$ HQMQ ratio \\
    \midrule
    hqmq\_s24\_r3  & 3.17 & \textbf{$5.05\times$} smaller \\
    hqmq\_s24\_r4  & 3.42 & $4.68\times$ \\
    hqmq\_s48\_r4  & 3.67 & $4.36\times$ \\
    hqmq\_s96\_r4  & 3.92 & $4.08\times$ \\
    hqmq\_s192\_r4 & 4.17 & $3.84\times$ \\
    hqmq\_s192\_r6 & 4.67 & $3.43\times$ \\
    \bottomrule
  \end{tabular*}
  \caption{Storage-format memory accounting per HQMQ config (per-element bits including the $16/d_h$ per-token scale).}
  \label{tab:memory}
\end{table}

\begin{figure}[h]
  \centering
  \includegraphics[width=\textwidth]{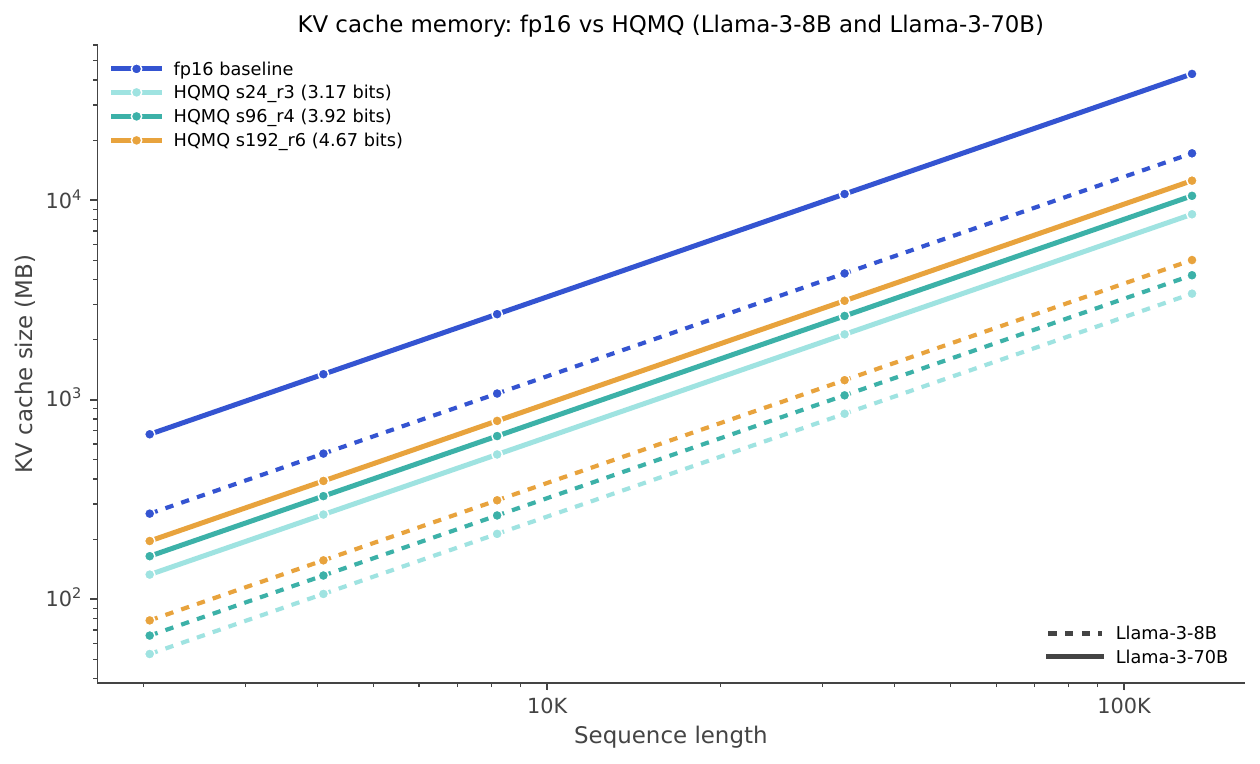}
  \caption{KV cache memory: fp16 (deep blue) vs HQMQ at three bit budgets for Llama-3-8B (dotted) and Llama-3-70B (solid). HQMQ s24\_r3 ($5.05\times$ compression) makes 70B / 128k-context inference fit on a single 24 GB consumer GPU.}
  \label{fig:memory}
\end{figure}

\section{K-chunk outlier diagnosis (Mistral vs Qwen2.5)}
\label{app:outlier-diag}

\begin{figure}[h]
  \centering
  \includegraphics[width=\textwidth]{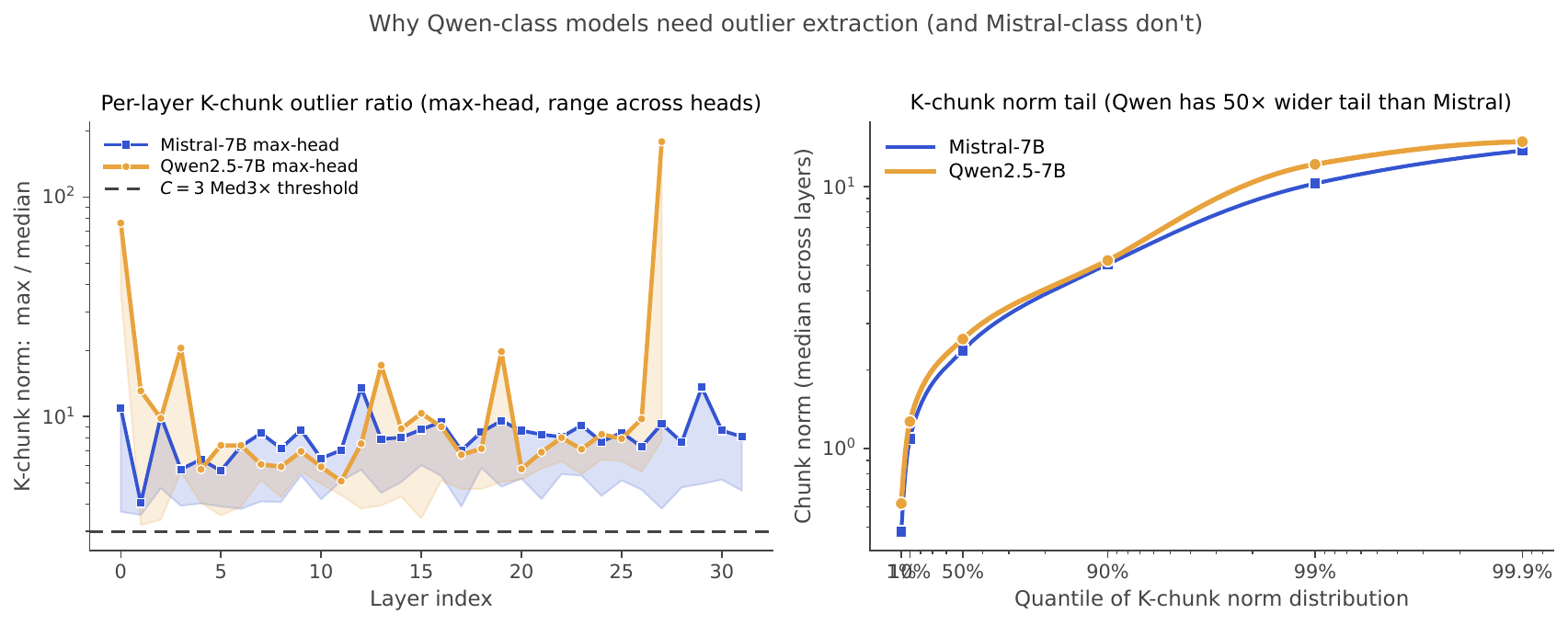}
  \caption{K-chunk outlier diagnosis on Mistral-7B vs Qwen2.5-7B. \emph{Left}: per-layer max-over-heads ratio of $K_{\max} / K_{\mathrm{med}}$. Qwen2.5 (amber) exceeds the Med3$\times$ threshold ($C{=}3$, dashed) in $\geq 95\%$ of layers, with peaks $>100\times$; Mistral (blue) hovers near $C{=}3$ with peaks $\lesssim 10\times$. \emph{Right}: K-chunk-norm quantile profile (median across layers). Qwen's 99.9th-percentile chunk norm is ${\sim}50\times$ its median; Mistral's is ${\sim}8\times$. The architectural difference in outlier statistics directly explains why Med3$\times$ extraction is essential for Qwen-class models and only marginally helpful for Mistral-class.}
  \label{fig:outlier-diag}
\end{figure}

\section{Outlier-multiplier sweep on Qwen2.5}
\label{app:outlier-sweep}

\begin{figure}[h]
  \centering
  \includegraphics[width=\textwidth]{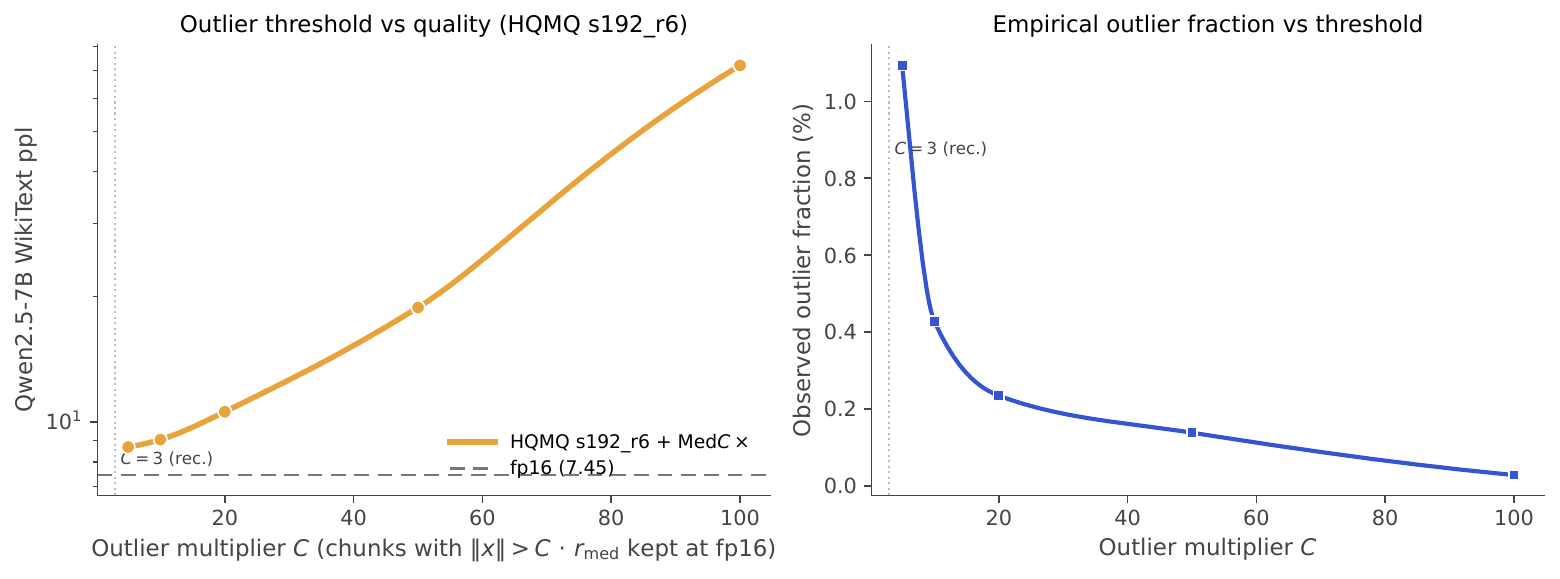}
  \caption{Outlier-multiplier sweep on Qwen2.5-7B (HQMQ s192\_r6). \emph{Left}: perplexity vs $C$. \emph{Right}: empirical outlier fraction vs $C$. Med3$\times$ extracts ${\sim}3\%$ of chunks and gives the best ppl; Med100$\times$ ($0.03\%$ extracted) is catastrophic.}
  \label{fig:outlier-sweep}
\end{figure}

\section{Qwen2.5-7B downstream task breakdown}
\label{app:qwen-downstream}

\begin{figure}[h]
  \centering
  \includegraphics[width=\textwidth]{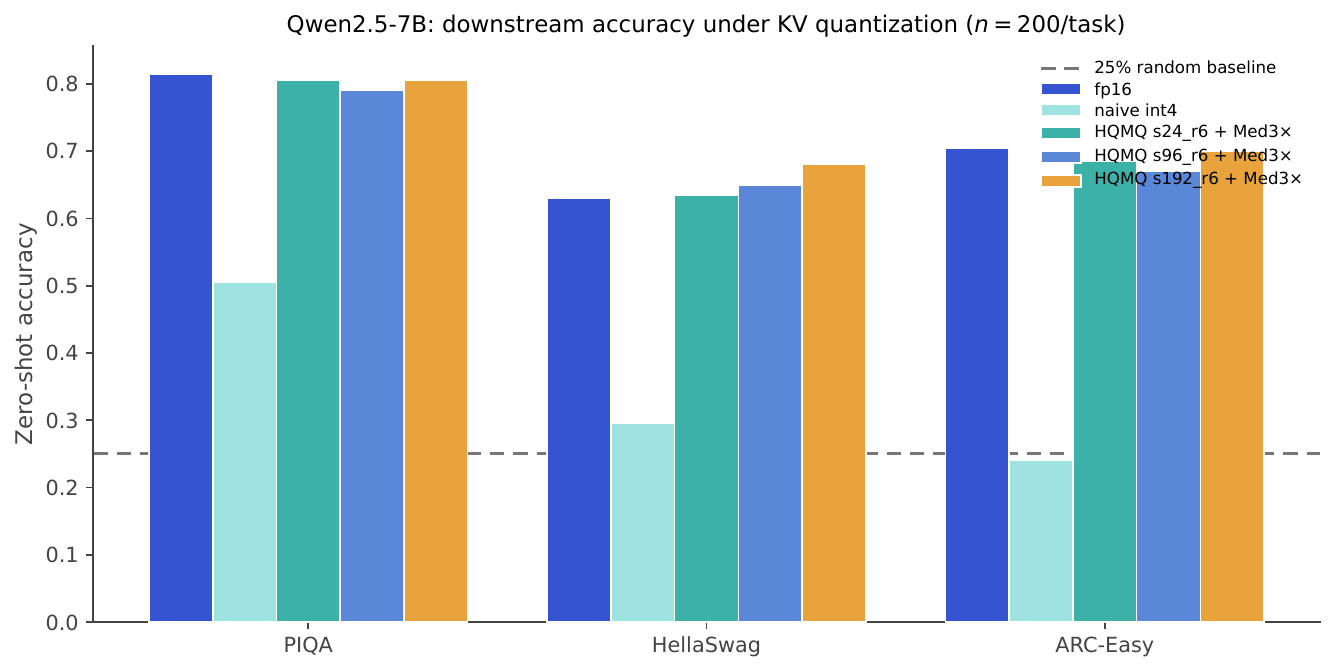}
  \caption{Qwen2.5-7B downstream task accuracy (PIQA / HellaSwag / ARC-Easy at $n{=}200$/task). Naive int4 collapses to (or below) the 25\% random baseline on HellaSwag and ARC, while the three HQMQ + Med3$\times$ configurations are within $\pm 2$ percentage points of fp16 on every task at $\sim 5$ bits. HQMQ s192\_r6 + Med3$\times$ exceeds fp16 on HellaSwag (0.680 vs 0.630), which we attribute to the slight regularizing effect of the radius quantizer at small-sample evaluation ($n{=}200$/task) rather than a real quality improvement. The same data is in the PIQA / HSwag / ARC-E columns of Table~\ref{tab:qwen}.}
  \label{fig:qwen-downstream}
\end{figure}

\section{gpt-oss-20b sparse-MoE results}
\label{app:gptoss}

gpt-oss-20b is OpenAI's open sparse-MoE release (24 layers, 8 KV heads, $d_h{=}64$ set explicitly via the \texttt{head\_dim} config field, ${\sim}20$B total / ${\sim}3.6$B active parameters in MXFP4). We loaded the model in its native MXFP4 weight format ($13.8$ GB on a 24~GB GPU)---the intended deployment format. The fp16 baseline of $446.8$ ppl on WikiText-103 is anomalously high (the model is instruction-tuned and the MXFP4 weights add baseline drift), but the \emph{relative} HQMQ pattern matches all other models we tested.

\begin{table}[h]
  \centering
  \small
  \begin{tabular*}{\textwidth}{@{\extracolsep{\fill}}lrrr@{}}
    \toprule
    config & bits & ppl & $\Delta$ vs fp16 \\
    \midrule
    fp16                              & 16.00 & 446.8  & 0 \\
    \textbf{hqmq\_s96\_r6 + Med3$\times$} & \textbf{5.25} & \textbf{460.4} & \textbf{+13.6} \\
    \textbf{hqmq\_s192\_r4}           & 4.04 & \textbf{469.5} & \textbf{+22.7} \\
    hqmq\_s24\_r6 + Med3$\times$      & 4.78 & 542.2 & +95.4 \\
    hqmq\_s96\_r4                     & 3.79 & 553.8 & +107.0 \\
    hqmq\_s48\_r4                     & 3.54 & 688.1 & +241.3 \\
    hqmq\_s24\_r3                     & 3.04 & 1{,}039 & +593 \\
    naive int4                        & 4.00 & 1{,}666 & +1{,}219 \\
    naive int3                        & 3.00 & 3{,}528 & +3{,}081 \\
    \bottomrule
  \end{tabular*}
  \caption{gpt-oss-20b WikiText-103 perplexity ($5w \times 2048$). HQMQ Pareto-dominates the naive baseline by $3$--$3.5\times$ at matched bits; HQMQ s96\_r6 + Med3$\times$ at 5.25 bits is within $3\%$ of fp16 ($\Delta\,{+}13.6$ of $446.8$). The MXFP4-quantized weights and instruction-tuning explain the absolute baseline; the relative pattern is consistent with the other five models.}
  \label{tab:gptoss-ppl}
\end{table}

HQMQ s96\_r6 + Med3$\times$ at 5.25 bits is within $3\%$ of fp16 relative ppl, with Med3$\times$ extracting ${\sim}2\%$ of chunks at $C{=}3$---consistent with the recipe transferring across all six tested models without retuning. The absolute fp16 ppl of 446.8 is an artifact of evaluating a sparse-MoE chat/reasoning model on a raw text-LM benchmark (WikiText-103); the relevant claim is the relative $\Delta$, not the absolute number. The Pareto pattern matches the dense models: HQMQ s192\_r4 at 4.04 bits is $3.5\times$ better than naive int4 at matched bits ($469.5$ vs $1666$); HQMQ s24\_r3 is $3.4\times$ better than naive int3 ($1039$ vs $3528$).

\section{Calibration-free ablation (full seed-variance table)}
\label{app:calib-free}

Five-seed variance of HQMQ on Mistral-7B across four codebook configurations. The coefficient of variation in end-task perplexity is $<0.15\%$ across all configs and all seeds---an order of magnitude tighter than typical evaluation noise on $20w \times 2048$, supporting the calibration-free claim in the main paper (Section~\ref{sec:abl-calib}).

\begin{table}[h]
  \centering
  \small
  \begin{tabular*}{\textwidth}{@{\extracolsep{\fill}}lrlrr@{}}
    \toprule
    config & bits & ppl range & std & CoV \\
    \midrule
    hqmq\_s24\_r3  & 3.04 & 5.759--5.781 & 0.008 & \textbf{0.14\%} \\
    hqmq\_s96\_r4  & 3.79 & 5.567--5.580 & 0.004 & \textbf{0.07\%} \\
    hqmq\_s192\_r4 & 4.04 & 5.538--5.551 & 0.005 & \textbf{0.09\%} \\
    hqmq\_s192\_r6 & 4.54 & 5.531--5.543 & 0.004 & \textbf{0.08\%} \\
    \bottomrule
  \end{tabular*}
  \caption{Seed variation of HQMQ on Mistral-7B ($20w \times 2048$, 5 seeds $\in \{0, 1, 7, 42, 1337\}$). Coefficient of variation is $<0.15\%$ across all configs and all seeds.}
  \label{tab:seed-variance}
\end{table}

\section{End-to-end attention latency}
\label{app:e2e-latency}

Per-step kernel latency only tells part of the story; the deployment-relevant question is total attention time over a realistic prefill $+$ generation workload. For a 4k-prompt $+$ 1024-token generation on Mistral-class GQA (RTX 4090, fp16):

\begin{table}[h]
  \centering
  \scriptsize
  \begin{tabular*}{\textwidth}{@{\extracolsep{\fill}}lrrr@{}}
    \toprule
    config & prefill & decode ($\times 1024$) & total \\
    \midrule
    fp16 SDPA (dense)        & $0.98$ ms & $14$ ms & $15$ ms \\
    fake-quant pipeline      & $1.19$ ms & $249$ ms & $250$ ms \\
    \textbf{Fused HQMQ}      & \textbf{$5.70$ ms} & \textbf{$34$ ms} & \textbf{$40$ ms} \\
    \bottomrule
  \end{tabular*}
  \caption{Total attention latency for prefill + $1024$-token generation at $T_{\text{prompt}}{=}4{,}096$. The fake-quant pipeline pays a $1.3\,\text{ms} \to 244\,\text{ms}$ per-step price after prefill; the fused kernel keeps decode cheap and dominates total time.}
  \label{tab:e2e-attn}
\end{table}

HQMQ + the fused kernel is $\mathbf{6.3\times}$ faster than the fake-quant pipeline and $\sim 2.7\times$ slower than the dense-fp16 FlashAttention upper bound. The gap to fp16 closes further as the generation tail grows (the fused kernel's decode-step latency is constant in $T_{kv}$ while the fake-quant pipeline grows linearly). For long-context batched serving---the production deployment HQMQ targets---HQMQ + the fused kernel makes long-context inference practical at $5\times$ less KV memory with $\sim 2.7\times$ end-to-end latency overhead, vs $16\times$ for the fake-quant baseline.

\section{Ablation: HQMQ vs uncalibrated additive VQ}
\label{app:additive-vq-ablation}

To isolate the contribution of the multiplicative quaternion structure (as opposed to ``having a large effective codebook''), we compare HQMQ against an additive VQ baseline of the form $c \approx c_1[i_1] + c_2[i_2]$---the CommVQ structure, but with \emph{random fixed} codebooks (no training). At matched per-element bit counts and codebook size on Mistral-7B:

\begin{table}[h]
  \centering
  \small
  \begin{tabular*}{\textwidth}{@{\extracolsep{\fill}}lrrrr@{}}
    \toprule
    target bits & HQMQ ppl ($b$) & Additive ppl ($b$) & ratio \\
    \midrule
    3.04 & \textbf{5.775} (3.04) & 71.861 (3.04) & \textbf{$12.4\times$ better} \\
    3.54 & \textbf{5.618} (3.54) & 33.642 (3.79) & \textbf{$6.0\times$ better} \\
    4.04 & \textbf{5.551} (4.04) & 16.522 (4.79) & \textbf{$3.0\times$ better} \\
    \bottomrule
  \end{tabular*}
  \caption{HQMQ vs uncalibrated additive VQ on Mistral-7B ($20w \times 2048$). The multiplicative composition is $3{-}12\times$ better at matched or fewer bits. Additive entries are at the natural bit count of $\lceil 2 \log_2 K \rceil + b_r$; the row at 3.04 bits is matched exactly, the other rows give the additive baseline \emph{more} bits and HQMQ still wins decisively.}
  \label{tab:abl-addvq}
\end{table}

The multiplicative composition over Hurwitz quaternions provides $3$--$12\times$ better quality than additive composition at matched (or fewer) bits when neither is trained. CommVQ~\cite{commvq2025} achieves 1-bit lossless quality by \emph{training} its codebooks; HQMQ's contribution is that it doesn't need that training step in the 3--5 bit regime, and that the multiplicative-group composition rule is the key ingredient making this possible. This ablation complements the Qwen2.5 disentanglement (main paper Section~\ref{sec:abl-disentangle}): the disentanglement shows the multiplicative codebook is necessary on outlier-heavy attention, while this table shows it is also more bit-efficient than the obvious alternative composition rule (additive) on outlier-free attention.

\section{Negative result: sensitivity-driven mixed precision}
\label{app:mixed-prec}

We tested whether per-layer mixed-precision allocation~\cite{kvtuner2025}---assigning more bits to layers measured to be more sensitive to KV quantization---could push HQMQ to lossless quality at a lower average bit count. We computed sensitivity by running an all-high HQMQ baseline (s192\_r6 = 4.54 bits everywhere), then downquantizing one layer at a time to s24\_r3 (3.04 bits) and recording the per-layer $\Delta$ppl. We then greedily allocated bits with a richer menu (s24\_r2 through s192\_r6) at target average bits $= 3.5$.

\begin{table}[h]
  \centering
  \scriptsize
  \begin{tabular*}{\textwidth}{@{\extracolsep{\fill}}lrrr@{}}
    \toprule
    config & avg bits & ppl & $\Delta$ vs fp16 \\
    \midrule
    fp16 & 16.00 & 6.081 & 0 \\
    HQMQ uniform s48\_r4 & 3.54 & \textbf{6.256} & \textbf{+0.175} \\
    HQMQ mixed-prec (sens.) & 3.50 & 6.553 & +0.471 \\
    \bottomrule
  \end{tabular*}
  \caption{Mixed-precision HQMQ negative result on Mistral-7B. The ``sens.''\ row uses a sensitivity-driven layer-wise bit allocation; including a sub-3-bit floor in the bit menu (s24\_r2) drags overall quality below uniform allocation.}
  \label{tab:mixed-prec}
\end{table}

The mixed allocation made 8 sensitive layers s192\_r6 (4.54 bits) and the rest s24\_r2 (2.79 bits). Per-layer error introduced at the s24\_r2 floor (the sub-3-bit regime where HQMQ breaks) dragged overall quality below uniform allocation. \textbf{Uniform allocation at moderate bits is sufficient and simpler.} Whether more careful mixed-precision schemes (e.g., excluding the broken floor, finer sensitivity probing) can recover this gap is left to future work.

\section{Llama-3-70B note}
\label{app:llama70b}

We did not run empirical perplexity / downstream evaluation on Llama-3-70B because the bf16 weights ($\sim$140 GB) do not fit on the single 24 GB RTX 4090 used in this work even with CPU offloading at usable throughput. The memory-accounting numbers we cite for Llama-3-70B (Table~\ref{tab:memory}, Fig.~\ref{fig:memory}) are \emph{analytical}: they follow directly from the bit/element rate and the model's KV-cache shape ($80$ layers, $8$ KV heads, $d_h = 128$). Empirical Llama-3-70B validation requires multi-GPU infrastructure beyond our setup; we expect the qualitative pattern observed on Llama-3-8B and Mistral-7B (HQMQ Pareto-dominant in 3--5 bits, Med3$\times$ outlier extraction a safe default) to transfer to 70B-scale Llama models, but the absence of an empirical check at 70B scale is a real limitation we explicitly flag.

\section{Additional results}

\subsection{Full Qwen2.5-7B outlier-multiplier sweep}

\begin{table}[h]
  \centering
  \small
  \begin{tabular*}{\textwidth}{@{\extracolsep{\fill}}lrrr@{}}
    \toprule
    config & bits & ppl & $\Delta$ vs fp16 \\
    \midrule
    fp16 & 16.00 & 7.590 & 0 \\
    s192\_r6 + Med3$\times$ & 5.15 & 8.830 & +1.241 \\
    s192\_r6 + Med4$\times$ & 5.00 & 8.999 & +1.409 \\
    s96\_r6 + Med3$\times$  & 4.91 & 9.026 & +1.436 \\
    s192\_r6 + Med5$\times$ & 4.91 & 9.259 & +1.669 \\
    s96\_r6 + Med5$\times$  & 4.67 & 9.362 & +1.772 \\
    s24\_r6 + Med3$\times$  & 4.42 & 9.692 & +2.102 \\
    s24\_r6 + Med5$\times$  & 4.17 & 10.394 & +2.804 \\
    s192\_r6 (no outlier)   & 4.54 & 109.0 & +101.4 \\
    s96\_r6 (no outlier)    & 4.29 & 98.4  & +90.8 \\
    naive int4              & 4.00 & 18{,}079 & +18{,}072 \\
    \bottomrule
  \end{tabular*}
  \caption{Qwen2.5-7B WikiText-103 perplexity at varying outlier multipliers $C$ and codebook sizes $S$. Without outlier extraction every HQMQ config exceeds 90 ppl; with $C{=}3$, all configurations stay in the 8--10 ppl range. Naive int4 remains broken regardless.}
  \label{tab:qwen-ppl-full}
\end{table}

\subsection{Full wall-clock latency table}

\begin{table}[h]
  \centering
  \small
  \begin{tabular*}{\textwidth}{@{\extracolsep{\fill}}lrrr@{}}
    \toprule
    config & bits & tok/s @ 1k & tok/s @ 4k \\
    \midrule
    fp16          & 16.00 & 39.10 (100\%) & 8.44 (100\%) \\
    naive int4    & 4.00  & 32.98 (84\%)  & 8.29 (98\%) \\
    hqmq\_s24\_r3 & 3.04  & 16.83 (43\%)  & 6.66 (79\%) \\
    hqmq\_s96\_r4 & 3.79  & 16.33 (42\%)  & 1.66 (20\%) \\
    hqmq\_s192\_r4 & 4.04 & 15.50 (40\%)  & 0.98 (12\%) \\
    \bottomrule
  \end{tabular*}
  \caption{Greedy-decode throughput on Mistral-7B (RTX 4090, bf16) under our fake-quant prototype. Pessimistic upper bound on overhead; a production fused kernel would eliminate most of the gap.}
  \label{tab:wallclock-app}
\end{table}

\subsection{Negative result: JL residual does not help}

Adding a Johnson--Lindenstrauss + 1-bit sign residual correction~\cite{qjl2025} on top of HQMQ does not help: at matched bits, additional codewords (larger $S$) yield more quality improvement than spending the same bits on a residual code. This contrasts with TurboQuant's PolarQuant + QJL composition, where the PolarQuant primary uses a smaller direction codebook and the JL residual is essential.

\subsection{Sub-3-bit regime}

HQMQ with $S = 24$, $b_r = 2$ (2.79 bits/element) produces unstable quality on all models tested (ppl rises with high seed variance below 3 bits). This is the regime where CommVQ~\cite{commvq2025} (trained codebooks designed to commute with RoPE) outperforms HQMQ. We do not target this regime; HQMQ is a 3--5 bit method.

\section{Additional model: Phi-3.5-mini}
\label{app:additional-models}

We tested one further architecture outside the five main-paper models: Microsoft's Phi-3.5-mini-instruct (full-MHA, 32 layers, 32 KV heads, $d_h{=}96$, available via \texttt{microsoft/Phi-3.5-mini-instruct}). Phi-3.5-mini gives a $4\times$ larger KV cache per token than the GQA main-paper models, so KV quantization matters more in absolute terms there. $d_h{=}96$ is divisible by 4 so HQMQ applies without padding.

\begin{table}[h]
  \centering
  \small
  \begin{tabular*}{\columnwidth}{@{\extracolsep{\fill}}lrrr@{}}
    \toprule
    config & bits & ppl & $\Delta$ vs fp16 \\
    \midrule
    fp16                    & 16.00 & 6.292   & 0 \\
    \textbf{hqmq\_s96\_r4}  & 3.79  & \textbf{6.537}  & \textbf{+0.245} \\
    \textbf{hqmq\_s48\_r4}  & 3.54  & \textbf{6.612}  & \textbf{+0.320} \\
    naive int4              & 4.00  & 6.765   & +0.473 \\
    hqmq\_s24\_r3           & 3.04  & 7.121   & +0.829 \\
    naive int3              & 3.00  & 12.922  & +6.630 \\
    \bottomrule
  \end{tabular*}
  \caption{WikiText-103 perplexity on Phi-3.5-mini ($20w \times 2048$). HQMQ s48\_r4 at 3.54 bits beats naive int4 at 4.00 bits; HQMQ s96\_r4 at 3.79 bits gives $\Delta\,{+}0.24$ ppl points from fp16. The Pareto pattern matches Mistral-7B and Llama-3-8B exactly.}
  \label{tab:additional-models}
\end{table}

The full gpt-oss-20b sparse-MoE sweep is in Appendix~\ref{app:gptoss}.

\subsection{Padding wrapper for $d_h$ not divisible by 4}
\label{app:padded-hqmq}

For models where $d_h \bmod 4 \neq 0$ (rare in practice but possible---e.g.\ if a future architecture chooses $d_h = 45$ via the implicit \texttt{hidden\_size / num\_attention\_heads} convention), we implement a padding wrapper that pads the head to the next multiple of 4 before HQMQ and truncates back on dequant. The bit overhead is $\lceil d_h / 4 \rceil \cdot 4 / d_h - 1$ (e.g.\ $48/45 - 1 = 6.7\%$ at $d_h = 45$). Source code: \texttt{src/quantizers/hqmq\_padded.py}. We did not need to exercise this wrapper on any of the six models tested in this paper, since all use $d_h \in \{64, 96, 128\}$.

\subsection{Bit accounting examples}

\begin{table}[h]
  \centering\small
  \begin{tabular*}{\textwidth}{@{\extracolsep{\fill}}lrrrrr@{}}
    \toprule
    config & $\log_2(24S)$ & $b_r$ & per-chunk bits & per-element bits & with $16/d_h$ \\
    \midrule
    s24\_r3   & 9.17  & 3 & 12.17 & 3.04 & 3.17 \\
    s24\_r4   & 9.17  & 4 & 13.17 & 3.29 & 3.42 \\
    s48\_r4   & 10.17 & 4 & 14.17 & 3.54 & 3.67 \\
    s96\_r4   & 11.17 & 4 & 15.17 & 3.79 & 3.92 \\
    s192\_r4  & 12.17 & 4 & 16.17 & 4.04 & 4.17 \\
    s192\_r6  & 12.17 & 6 & 18.17 & 4.54 & 4.67 \\
    \bottomrule
  \end{tabular*}
  \caption{Worked bit accounting for representative HQMQ configs at $d_h = 128$.}
\end{table}

\end{document}